\documentclass[journal]{IEEEtran}
\usepackage{amsmath,amsfonts}
\newcommand{\Rmnum}[1]{\uppercase\expandafter{\romannumeral #1}}
 
\usepackage{algorithmic}
\usepackage{makecell}
\usepackage{array}
\usepackage[caption=false,font=normalsize,labelfont=sf,textfont=sf]{subfig}
\newtheorem{remark}{Remark}
\usepackage{textcomp}
\usepackage{stfloats}
\usepackage{url}
\usepackage{verbatim}
\usepackage{graphicx}
\usepackage{hyperref}  
\usepackage{booktabs}
\usepackage{multirow}
\usepackage{colortbl} 
\usepackage[numbers,sort&compress]{natbib}
\usepackage[table,xcdraw]{xcolor}

\def\BibTeX{{\rm B\kern-.05em{\sc i\kern-.025em b}\kern-.08em
    T\kern-.1667em\lower.7ex\hbox{E}\kern-.125emX}}
\usepackage{balance}
\begin{document}
\title{\huge Efficient Spike-driven Transformer for High-performance Drone-View Geo-Localization} 
\author{
\thanks{}}
\author{Zhongwei Chen, \textit{Student Member, IEEE}, Hai-Jun Rong \textit{Senior Member, IEEE}, Zhao-Xu Yang, \textit{Member, IEEE}, Guoqi Li, \textit{Member, IEEE}\\
\thanks{This paper is submitted for review on December 07, 2026. This work was supported in part by the Key Research and Development Program of Shaanxi, PR China (No. 2023-YBGY-235), the National Natural Science Foundation of China (No. 61976172 and No. 12002254), Major Scientific and Technological Innovation Project of Xianyang, PR China (No. L2023-ZDKJ-JSGG-GY-018), and Shanghai NeuHelium Neuromorphic Technology Co., Ltd. Zhongwei Chen and Hai-Jun Rong contributed equally to this work. (Corresponding authors: Zhao-Xu Yang and Guoqi Li)}

\thanks{Zhongwei Chen, Zhao-Xu Yang and Hai-Jun Rong  are with the State Key Laboratory for Strength and Vibration of Mechanical Structures, Shaanxi Key Laboratory of Environment and Control for Flight Vehicle, School of Aerospace Engineering, Xi’an Jiaotong University, Xi’an 710049, PR China (e-mail:ISChenawei@stu.xjtu.edu.cn; yangzhx@xjtu.edu.cn; hjrong@mail.xjtu.edu.cn).  

Guoqi Li is with the Institute of Automation, Chinese Academy of Sciences, Beijing 100190, China, also with the School of Artificial Intelligence, University of Chinese Academy of Sciences, Beijing 101408, China, and
also with the Peng Cheng Laboratory, Shenzhen 518000, China (e-mail: guoqi.li@ia.ac.cn).}}

\maketitle

\begin{abstract}
Traditional drone-view geo-localization (DVGL) methods based on artificial neural networks (ANNs) have achieved remarkable performance. However, ANNs rely on dense computation, which results in high power consumption. In contrast, spiking neural networks (SNNs), which benefit from spike-driven computation, inherently provide low power consumption. Regrettably, the potential of SNNs for DVGL has yet to be thoroughly investigated. Meanwhile, the inherent sparsity of spike-driven computation for representation learning scenarios also results in loss of critical information and difficulties in learning long-range dependencies when aligning heterogeneous visual data sources. To address these, we propose SpikeViMFormer, the first SNN framework designed for DVGL. In this framework, a lightweight spike-driven transformer backbone is adopted to extract coarse-grained features. To mitigate the loss of critical information, the spike-driven selective attention (SSA) block is designed, which uses a spike-driven gating mechanism to achieve selective feature enhancement and highlight discriminative regions. Furthermore, a spike-driven hybrid state space (SHS) block is introduced to learn long-range dependencies using a hybrid state space. Moreover, only the backbone is utilized during the inference stage to reduce computational cost.  To ensure backbone effectiveness, a novel hierarchical re-ranking alignment learning (HRAL) strategy is proposed. It refines features via neighborhood re-ranking and maintains cross-batch consistency to directly optimize the backbone. Experimental results demonstrate that SpikeViMFormer outperforms state-of-the-art SNNs. Compared with advanced ANNs, it also achieves a 13.24$\times$ reduction in the energy consumption of the inference stage and an 8.4$\times$ reduction in parameter count, with only a minimal drop in performance. {Our code is available at {\href{https://github.com/ISChenawei/SpikeViMFormer}{https://github.com/ISChenawei/SpikeViMFormer}}}
\end{abstract}


\section{Introduction}\label{intrduction}

\IEEEPARstart{D}{rone-View} geo-localization (DVGL) task aims to achieve accurate localization for drones in GNSS-denied environments by retrieving geo-referenced satellite-view images that best match drone-view images \cite{dai2021transformer,tian2021uav,zheng2020university,zhu2023sues}. With the widespread deployment of drones in disaster rescue \cite{Liu_Multi}, urban monitoring \cite{qin2025must}, and intelligent transportation, DVGL, which not only offers basic localization functions but also supports a robust backup localization capability for complex environments, has attracted increasing attention from both academia and industry. This task remains highly challenging and requires effective intra-view representation learning and cross-view alignment to address substantial discrepancies between drone and satellite imagery \cite{chen2024multi}. Fortunately, artificial neural networks (ANNs) have driven notable advances in DVGL and substantially improved retrieval accuracy \cite{du2024ccr,xia2024enhancing}.

\begin{figure}[t]
  \centering
  \includegraphics[width=3.4in]{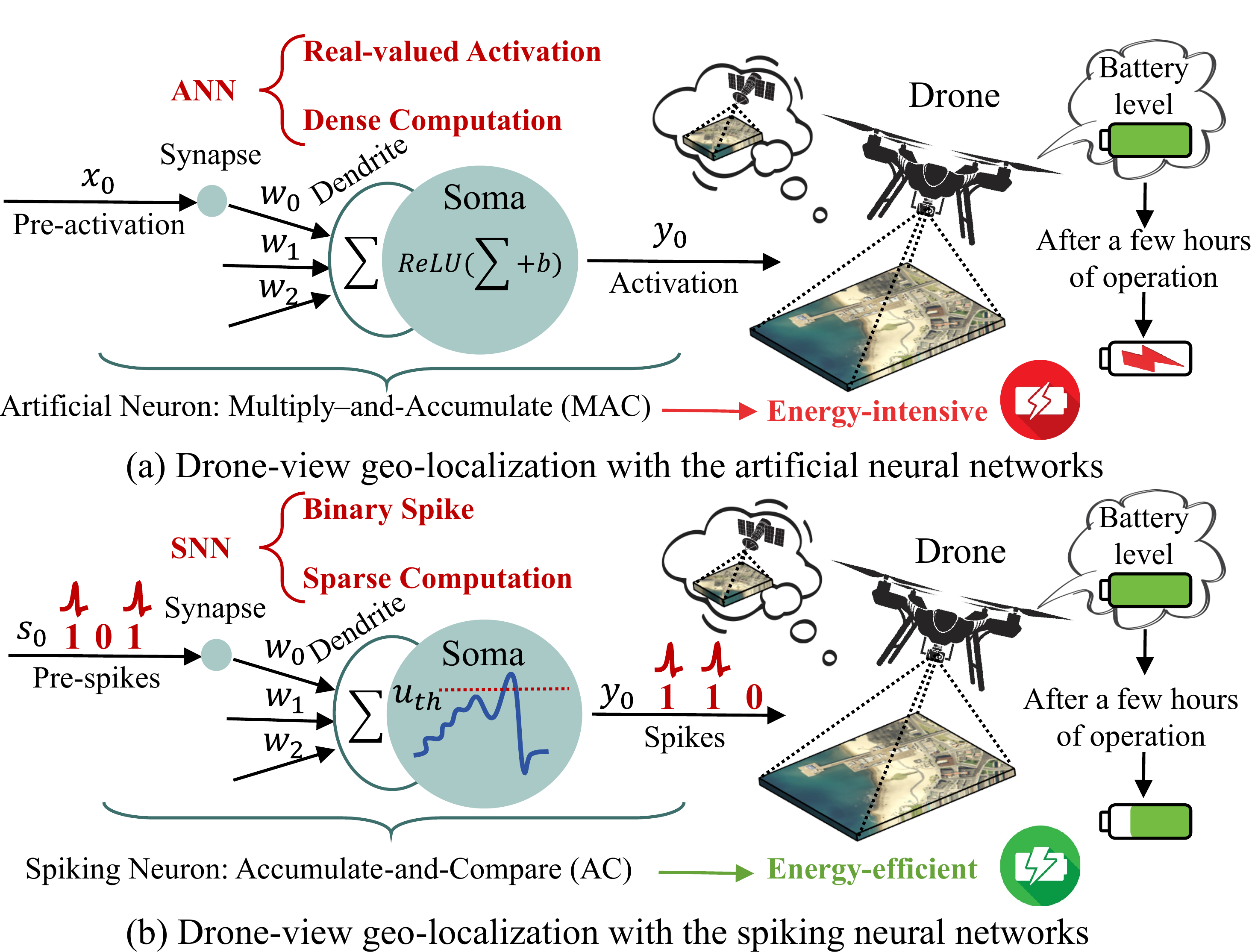}
  \caption{\textbf{DVGL with ANNs vs. SNNs}. (a) In ANN-based DVGL, neurons employ real-valued activations and dense MAC operations, which results in high activation density in convolution and attention layers. Such dense computation is energy-intensive, particularly on resource-constrained drone platforms, which leads to rapid battery depletion.
  (b) In contrast, SNN-based DVGL adopts an event-driven paradigm where neurons emit binary spikes only upon reaching the firing threshold. This replaces MAC with low-power AC operations and leverages sparse activations, thus achieving remarkable energy efficiency.}
  \label{fig1}
\end{figure}

\begin{figure*}[t]
  \centering
  \includegraphics[width=7.3in]{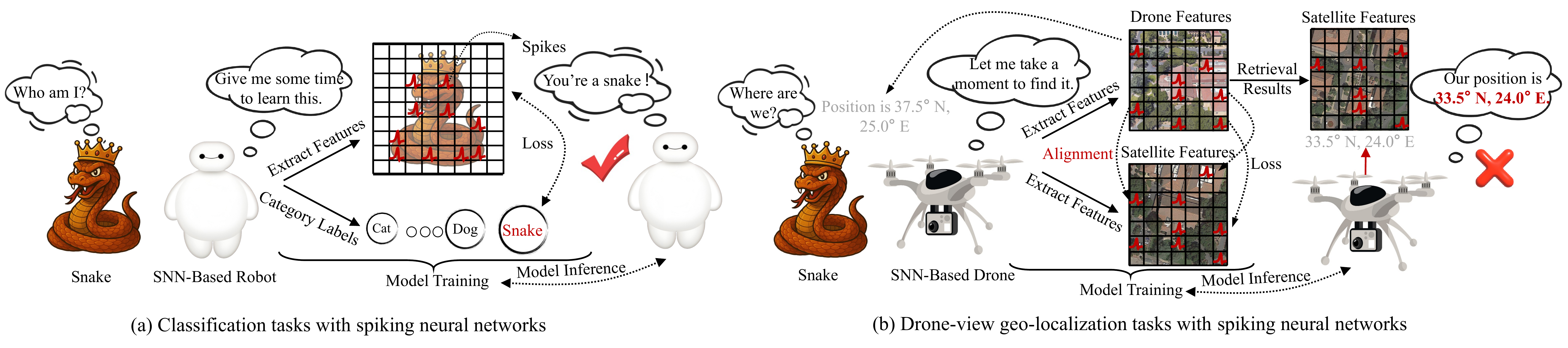}
  \caption{\textbf{Classification vs. DVGL with SNNs.} (a) In classification, SNNs only need to map an entire image to a class label, relying mainly on global discriminative cues such as textures or contours. Information compression and sparsity thus have little impact. (b) In DVGL, drone images from oblique viewpoints must be aligned with satellite images from top-down views. This cross-view scenario demands precise alignment. However, spike-induced sparsity amplifies alignment errors, making SNNs prone to confusion when visually similar but semantically different regions (e.g., rooftops with the same color) appear. Therefore, it is essential to capture and model long-range dependencies, as global contextual information helps disambiguate visually similar but semantically different regions and thus mitigates alignment errors.}
  \label{fig2}
\end{figure*}

Currently, the low-energy solution for resource-constrained drones has attracted our interest due to its potential for practical deployment. As shown in Fig.\ref{fig1}(a), most of the ANN-based methods in the DVGL task \cite{chen2024sdpl,ge2024multibranch,chen2025without} perform dense real-valued activations and multiply-and-accumulate (MAC) operations during the inference stage and maintain a high activation density in the convolution and attention layers. Such dense computations incur an inevitable substantial energy overhead. In particular, existing methods often utilize deep backbones \cite{liu2022convnet} and design high-complexity attention \cite{liu2022swin}, which further increases energy costs. In contrast, as shown in Fig.\ref{fig1}(b), spiking neural networks (SNNs) \cite{pei2019towards} adopt an event-driven paradigm that emits binary spikes only when the membrane potential of a neuron crosses a threshold. Here, MAC operations are replaced by lower-power accumulate-and-compare (AC) operations, and activations are highly sparse in both time and space \cite{yao2023attention,yao2023spike}. These properties can be translated into notable hardware-level energy efficiency \cite{deng2020model}. Consequently, the SNN-based solution for the DVGL task presents the possibility of meeting the dual demands of high accuracy and low energy consumption.

Despite the unique advantages of SNNs in energy efficiency compared with mature ANNs, SNNs still exhibit performance gaps in the complex visual tasks \cite{tavanaei2019deep}. 
Notably, some research efforts have been narrowing the performance gaps between SNNs and ANNs using surrogate gradient learning, ANN-to-SNN conversion, and architectural optimization \cite{meng2023towards}. In the classification task, for instance, state-of-the-art (SOTA) SNNs have already achieved competitive performance on mainstream benchmarks such as ImageNet \cite{deng2009imagenet} to close the performance of ANNs \cite{yao2024spikedriven,10848017}. More importantly, the emergence of neuromorphic chips further amplifies the hardware-level advantages of SNNs \cite{merolla2014million,davies2018loihi}. For example, under typical visual workloads, the asynchronous sensing-computing neuromorphic system on chip (SoC) Speck \cite{yao2024spike} achieves an operating power as low as 0.7 mW, which provides strong evidence of its efficiency in hardware deployment. These advances provide a practical groundwork for the DVGL task.

However, DVGL is more challenging than the existing classification task. As shown in Fig. \ref{fig2}(a), classification follows a relatively simple input-output mapping, where global discriminative cues such as textures and contours are sufficient, and the compression or sparsity caused by spiking neurons has little effect. In contrast, the direct application of SNNs to DVGL often suffers from poor performance. As illustrated in Fig.~\ref{fig2}(b), spike-induced sparsity often loses critical information in drone-view and satellite-view images. This weakens the intra-view discriminability. Moreover, SNNs tend to rely heavily on salient cues induced by spikes and hardly capture long-range dependencies. It makes SNNs prone to confusion from visually similar yet semantically different regions, such as rooftops of similar colors appearing in distinct geographic contexts. Therefore, existing SOTA SNNs still perform with limited capacity in the DVGL task.

In this work, SpikeViMFormer is proposed as the first high-performance SNN framework for the DVGL task. A lightweight spike-driven transformer backbone is adopted to extract features. Then, to mitigate the loss of critical information caused by spike-induced sparsity, the spike-driven selective attention (SSA) block is designed to employ a spike-driven gating mechanism to enforce selective information transmission and to emphasize discriminative regions. For better cross-view alignment, the spike-driven hybrid state space (SHS) block is introduced, which can obtain long-range dependencies by leveraging a hybrid state space. To preserve efficiency in resource-constrained deployment scenarios, both auxiliary blocks will be pruned during the inference stage. Consequently, a hierarchical re-ranking alignment learning (HRAL) strategy is proposed to ensure the effectiveness of the backbone. It refines features via neighborhood re-ranking and maintains cross-batch consistency to directly optimize the backbone. Extensive experiments demonstrate that SpikeViMFormer surpasses SOTA SNNs and achieves competitive performance compared with advanced ANNs.

\begin{itemize}
  \item \textbf{Analysis of SNNs for DVGL Task}. We conduct a detailed analysis of the potential and challenges associated with applying SNN to the DVGL task and propose the first SNN framework SpikeViMFormer for DVGL.
  \item \textbf{Efficient Framework Design}. The SpikeViMFormer integrates SSA and SHS blocks to alleviate the loss of crucial information and difficulties in learning long-range dependencies caused by the inherent sparsity of SNNs. Meanwhile, an HRAL strategy is proposed as an active supervisory signal to maintain the effectiveness of the backbone and improve cross-view alignment.
  \item \textbf{SSA and SHS Blocks}. The SSA block uses spike-driven gates for local and global modulation to achieve selective feature enhancement. Meanwhile, the SHS block integrates local spatial priors with the hybrid state space module by alternately reshaping features between sequence and spatial formats to learn long-range dependencies and local spatial representations. The SSA block serves as the critical informational foundation for the SHS block, while the SHS block builds upon this foundation to further learn contextual representations.
  \item \textbf{HRAL Strategy}. We propose a novel HRAL strategy that recasts conventional re-ranking from an inference heuristic into a supervised training strategy. It learns reliable cross-batch neighbor information by integrating $k$-reciprocal consistency, Gaussian weighting, and query-expansion smoothing to obtain refined features. It further guides the backbone features toward consistency with their refined counterparts and directly facilitates cross-view alignment.
  \item \textbf{Performance}. SpikeViMFormer achieves performance improvements in the DVGL task and maintains low energy consumption, which demonstrates the potential of SNNs in DVGL. It outperforms SOTA SNNs and achieves a 13.24$\times$ reduction in  the energy consumption of the inference stage and an 8.4$\times$ reduction in the parameter count.
\end{itemize}

The remainder of this paper is organized as follows. Section \ref{related works} systematically reviews previous research. In Section \ref{method}, the proposed SpikeViMFormer is presented in detail. The experimental results are reported and analyzed in Section \ref{results}. Finally, the conclusions are outlined in Section \ref{conclusions}.

\section{Related Works}\label{related works}

\subsection{Dross-View Geo-localization and its ANN-based Solutions}

DVGL is a sub-task of cross-view geo-localization \cite{wilson2021visual} that focuses on retrieving geo-referenced satellite images given query images taken by drones. DVGL is particularly important in GNSS-denied environments, where it provides a vision-based alternative for reliable localization. Recently, two dominant technologies have been developed as the foundational support for DVGL. One focuses on representation learning. LCM \cite{ding2020practical} mapped drone and satellite images into a shared feature space and formulated the task as a global classification problem to capture holistic scene structures. LPN \cite{wang2021each} proposed a ring-based partition strategy that split images into distance-aware regions and thus captured fine-grained contextual information around targets. IFSs \cite{ge2024multibranch} employed multi-branch designs to jointly leverage global and local features, while MFJR \cite{ge2024multi} introduced adaptive region suppression to filter irrelevant information and focus on salient targets. TransFG \cite{zhao2024transfg} employed transformer-based feature aggregation together with gradient-guided modules to combine global and local cues more effectively. For the other cross-view alignment technology, PCL \cite{tian2021uav} applied viewpoint projection transformations to mitigate spatial misalignment, followed by CGAN-based \cite{mirza2014conditional} style translation to bridge imaging gaps. CAMP \cite{wu2024camp} further leveraged contrastive attribute mining with position-aware partitioning to improve robustness when dealing with various perspectives and scales. DAC \cite{xia2024enhancing} enforced domain alignment and scene-consistency, constrained to refine the feature correspondence. More recently, MEAN \cite{chen2024multi} proposed a progressive multilevel enhancement strategy, global-to-local associations, and cross-domain alignment to exploit features from different levels.

It is worth noting that, although the above methods have achieved impressive performance, they all rely on ANN frameworks. Such frameworks often incur substantial energy overhead during deployment, as discussed in Section \ref{intrduction}. In addition, many methods employ deep backbones and complex attention mechanisms, which further increase energy consumption and computational cost. Unlike existing methods, the proposed SpikeViMFormer is built on an SNN framework, which effectively reduces energy consumption during the inference stage for the resource-constrained deployment scenario. Additionally, in the SpikeViMFormer, the auxiliary blocks are pruned during the inference stage, and only a lightweight backbone is used for the inference stage, which improves deployment feasibility and energy efficiency.

\subsection{Spiking Neural Networks and their Potential}
SNNs draw inspiration from biological neural systems and take advantage of spatiotemporal dynamics, spike-based encoding, and event-driven computation. They are increasingly recognized as a promising low-power, brain-inspired paradigm toward general artificial intelligence \cite{schuman2022opportunities}. Early studies focused on biological plausibility and neuroscience modeling \cite{maass1997networks}, while recent researches have been oriented towards computer vision applications \cite{yao2021temporal}. However, the non-differentiable nature of spiking mechanisms makes conventional backpropagation inapplicable to SNNs. Traditional methods relied mainly on biologically inspired unsupervised methods, such as the STDP algorithm \cite{bi2001synaptic}, which showed limited performance on complex tasks and could not scale to large datasets. To address this issue, two main methods for large-scale SNN training have been adopted, which were ANN-to-SNN \cite{hu2023fast} and direct training \cite{yao2023attention}. The former mapped pretrained ANN weights into the spiking domain for efficient learning, while the latter discretized the nonlinear dynamics of spiking neurons and employed spatiotemporal backpropagation with surrogate gradients \cite{eshraghian2023training}. These methods have greatly advanced the scalability of SNNs and enabled a series of promising models that demonstrated competitive performance close to ANNs on the tasks such as image classification \cite{10848017}, object detection \cite{luo2024integer} and semantic segmentation \cite{lei2025spike2former}. Additionally, the advent of neuromorphic chips has further accelerated the development of SNNs. Representative systems, such as Neurogrid \cite{benjamin2014neurogrid}, Tianjic \cite{pei2019towards}, and SpiNNaker \cite{hoppner2021spinnaker}, exploited the event-driven and sparse computation of SNNs to achieve substantial energy savings. This highlights their potential for the DVGL task on resource-constrained platforms.

Although those works have achieved remarkable progress in SNNs, most SOTA SNNs remain confined to the basic vision tasks. In more challenging scenarios such as DVGL, the inherent binarization and sparsity of spiking neurons still result in a substantial performance gap compared with ANNs. The proposed SpikeViMFormer is the first SNN framework designed for DVGL to achieve high performance with fewer parameter count and lower energy consumption in this task.

\section{SpikeViMFormer Framework}\label{method}
\textit{Problem Formulation}. In the DVGL task, we are given a dataset of paired drone-view and satellite-view images $\{I^i_d,I^j_s\}$, where $I^i_d$ denotes the $i$-th drone-view image with $N_d$ samples, and $I^j_s$ denotes the $j$-th satellite-view image with $N_s$ samples. The indices $i$ and $j$ are independent. The objective is to learn a shared feature space where matching pairs are pulled closer and non-matching pairs are pushed apart \cite{chen2024sdpl,xia2024enhancing}.

\subsection{Spiking Neuron}\label{Spiking Neuron}
To better understand the foundation of SNNs, the Leaky Integrate-and-Fire (LIF) neuron \cite{maass1997networks} is first introduced, which is the most widely used in SNN construction. The LIF neuron with a soft reset can be described as 
\begin{equation}
U[t] = H[t-1] + X[t],
\end{equation}
\begin{equation}\label{eq2}
S[t] = \Theta(U[t] - V_{{th}}),
\end{equation}
\begin{equation}
H[t] = \beta (U[t] -S[t]),
\end{equation}
where $t$ denotes the discrete timesteps. The membrane potential $U[t]$ integrates the previous state $H[t-1]$ with the current input $X[t]$. $S[t]$ is the output spike. $\Theta(\cdot)$ denotes the Heaviside step function, where $\Theta(x)=1$ if $x\ge0$, that is to say, a spike is generated once $U[t]$ exceeds the firing threshold $V_{\text{th}}$. $H[t]$ represents the post-firing membrane potential. Subsequent to the action potential firing, the membrane potential decays to $H[t]$ with a decay factor $\beta$.

However, conventional LIF neuron using Eq.(\ref{eq2}) inevitably introduces quantization errors when converting membrane potentials into binary spikes. Recent work \cite{luo2024integer} has proposed an Integer LIF (I-LIF) neuron to improve the numerical representation in the object detection task, but its extension to complex architectures often leads to gradient instability. Therefore, a Normalized Integer LIF (NI-LIF) neuron \cite{lei2025spike2former} has been proposed to preserve numerical representation and enhance training stability by normalization. The NI-LIF neuron is described as 
\begin{equation}
U[t] = H[t-1] + X[t],
\end{equation}
\begin{equation}\label{eq5}
S[t] = \text{Clip}\big(\text{round}(U[t]), 0, D\big) / D,
\end{equation}
\begin{equation}
H[t] = \beta \big(U[t] - S[t] \times D\big),
\end{equation}
where $\text{round}(\cdot)$ performs integer quantization of the membrane potential. 
$\text{Clip}(\cdot,0,D)$ restricts the quantized value within the range $[0,D]$ to avoid instability. $D$ denotes the upper bound of the integer activation, which serves as a normalization factor. By dividing the clipped integer quantization by $D$, the spike output $S[t]$ is scaled to $[0,1]$. In this work, the NI-LIF neuron is adopted by default unless otherwise specified. Moreover, for simplicity, the spiking neuron layer is denoted as $SN(\cdot)$.

\begin{figure*}[t]
  \centering
  \includegraphics[width=7.0in]{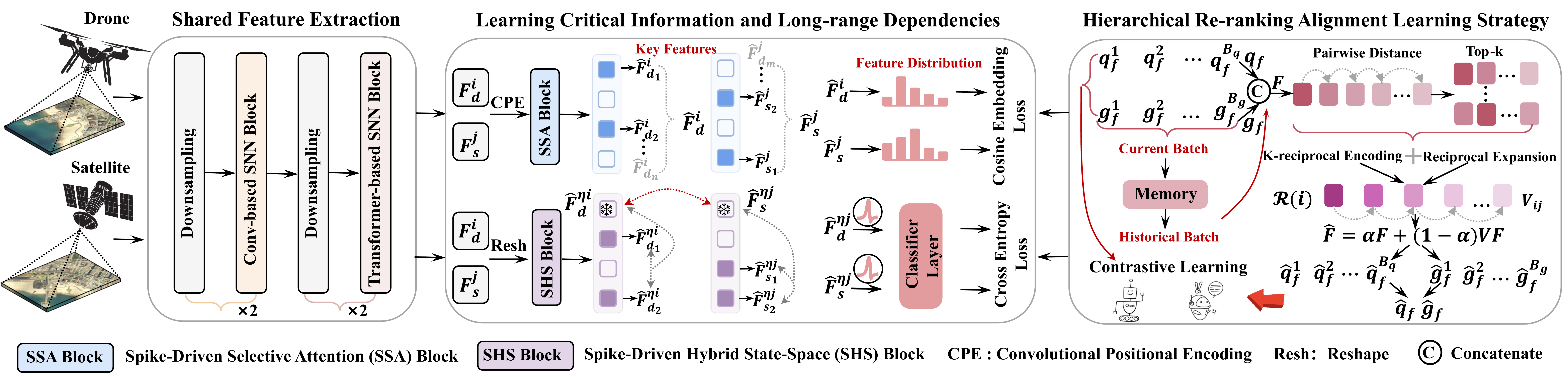}
  \caption{\textbf{Pipeline Overview}. The proposed SpikeViMFormer first adopts a dual-stream lightweight spike-driven transformer backbone with shared weights to extract coarse-grained features. On this basis, two auxiliary blocks are employed, including the SSA block and the SHS block. The SSA block is designed to selectively enhance and focus attention on discriminative regions. In addition, it leverages a cosine embedding loss to guide the narrowing of the representational gap between drone-view and satellite-view. Meanwhile, the SHS block is employed to capture long-range dependencies by leveraging hybrid state space dynamics, and it is optimized with a cross-entropy loss to improve semantic consistency and enhance cross-view alignment. In addition, the hierarchical re-ranking alignment strategy optimizes the backbone by enforcing cross-batch consistency, which ensures its effectiveness when auxiliary blocks are discarded at inference stage.}
  \label{fig3}
  \end{figure*}

\subsection{Spike-Driven Transformer Backbone}
In this work, a lightweight spike-driven transformer proposed in \cite{10848017} is used as the backbone, in which the classification head is pruned and only the feature extractor remains. As shown in Fig.~\ref{fig3}, the backbone contains two Conv-based SNN blocks and two Transformer-based SNN blocks, which are interleaved with Downsampling. The input image size is fixed at $384 \times 384$, and the weight-shared backbone is employed to extract common features from drone-view and satellite-view images. This process can be defined as
\begin{equation}
F^i_d = \mathcal{B}_{\theta}(I^i_d),
F^j_s = \mathcal{B}_{\theta}(I^j_s),
\end{equation}
where $\mathcal{B}_{\theta}$ denotes the weight-shared spike-driven transformer backbone. $F^i_d \in \mathbb{R}^{B \times C \times L}$ and $F^j_s \in \mathbb{R}^{B \times C \times L}$ are the extracted features from the drone-view image $I^i_d$ and the satellite-view image $I^j_s$, respectively. $B$ denotes the batch size, and $C$ is the number of feature channels. $L = H \times W$, where $H$ and $W$ represent the height and width of the feature, respectively.

\subsection{Spike-Driven Selective Attention (SSA) Block}
Although the backbone is capable of feature extraction, the sparse activation of spiking neurons easily causes critical information loss and decreased discriminability. To address this issue, the SSA block is designed to refine the features and enhance their discriminability. The backbone outputs features of the form $F^i_d \in \mathbb{R}^{B \times C \times L}$ and $F^j_s \in \mathbb{R}^{B \times C \times L}$ , which are flattened features of the original images. As a result, they contain global information but lack explicit local spatial and positional information. Therefore, before applying the SSA block, a convolutional positional encoding (CPE) \cite{chu2021conditional} is applied to learn local spatial relations and positional priors. For clarity, the processing of drone-view features is taken as an example to describe the CPE and the SSA block.

In CPE, $F^i_d$ is reshaped into a grid with spatial dimensions $P\times P$, where $P=\sqrt{L}$. Each cell within this grid is represented by an embedding that has $C$ channels. Therefore, $F^i_d$ transforms from a 1D sequence into a 2D feature. Then, a lightweight depthwise convolution accompanied by a residual connection is employed, in which the convolution operates in the local neighborhood of each position to generate a position-aware feature representation. Subsequently, the output undergoes a flattening operation, denoted as $Fla(\cdot)$, which reshapes it back into a 1D sequence, and this sequence is then fed into the SSA block. The CPE can be formulated as
\begin{equation}\label{eq8}
F_d^{i+}= F^{i}_d + \, Fla\left(\text{DWConv}\left(Res(F^{i}_d,P,P)\right)\right),
\end{equation}
where $\text{DWConv}(\cdot)$ is the depthwise convolution operator. $Res(\cdot, P, P)$ is the reshape operator that reconstructs the 1D sequence into a 2D feature of spatial dimensions $P \times P$.

\begin{figure}[t]
  \centering
  \includegraphics[width=3.3in]{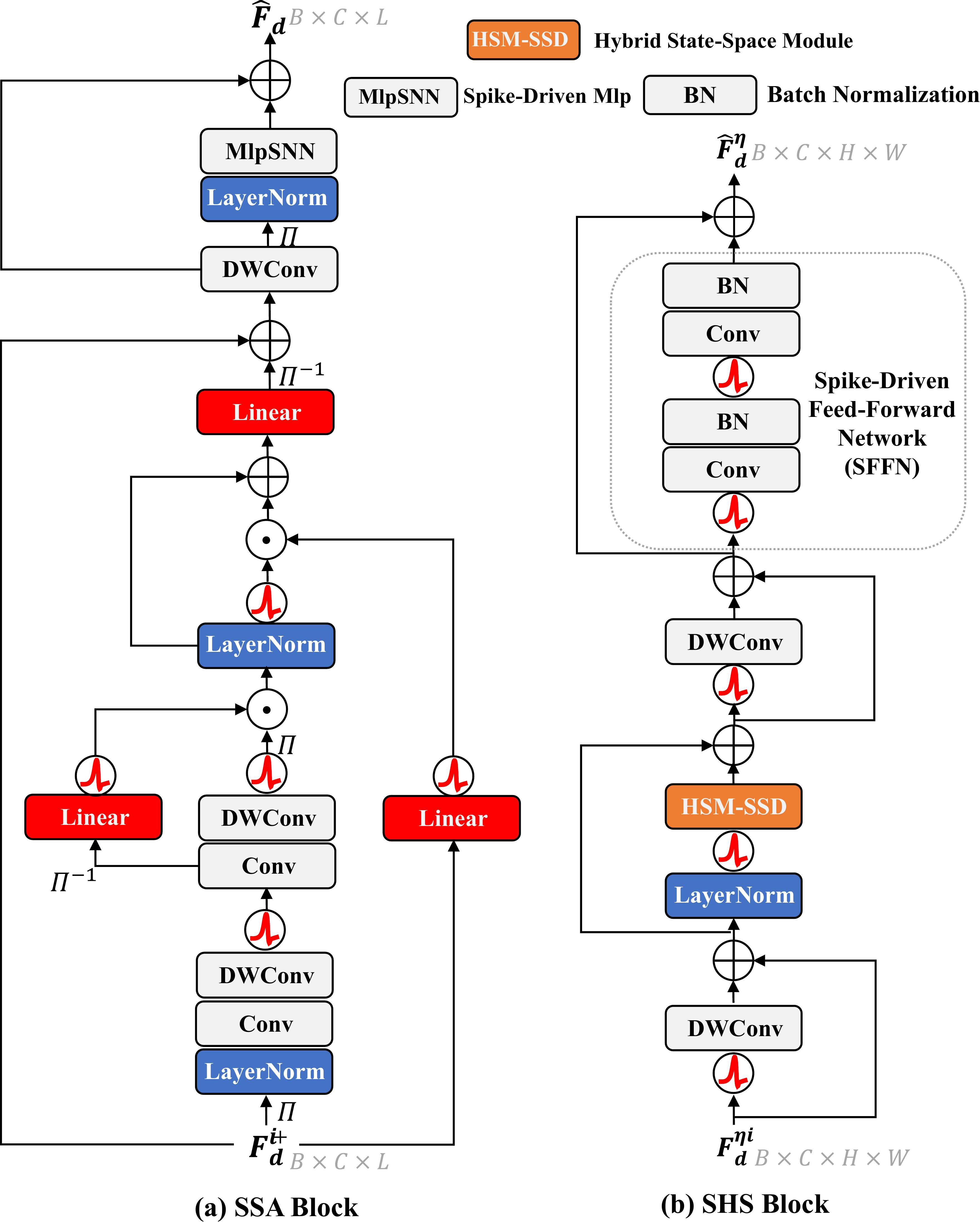}
  \caption{\textbf{SSA Block and SHS Block}. Illustration of (left) spike-driven selective attention (SSA) block and (right) spike-driven hybrid state space (SHS) block.}
  \label{fig4}
\end{figure}

In the SSA block, as illustrated in Fig.~\ref{fig4}, LayerNorm is applied first to stabilize the distribution $F_d^{i+}$. Then, a convolution followed by a depthwise convolution is performed, and the resulting output is processed by spike activation to obtain the refined feature ${\tilde{F}_d^{i+}}$, which is described as
\begin{equation}
\tilde{F}_d^{i+} = SN\left(\text{DWConv}\left(\text{Conv}\left(\mathrm{LN}\left(\Pi(F_d^{i+})\right)\right)\right)\right),
\end{equation}
where $\Pi(\cdot)$ denotes the permutation from $(B,C,L)$ to $(B,L,C)$, and $\Pi^{-1}(\cdot)$ denotes the inverse permutation from $(B,L,C)$ back to $(B,C,L)$. $\mathrm{LN}(\cdot)$ denotes LayerNorm. Meanwhile, a linear projection is used to generate a spike-driven gate $A$,
\begin{equation}
A = SN({F_d^{i+}} W_a),
\end{equation}
where $W_a$ is a learnable projection matrix. 

The feature ${\tilde{F}_d^{i+}}$ is convolved to form $Q$, and the following spiking branch performs depthwise convolution and spike activation to obtain the local feature $Q'$. In parallel, a linear  projection generates a spike-driven gate $G$. The fused representation $Q''$ is then obtained by $Q'$ with $G$ via element-wise multiplication, followed by layer normalization. This process can be described as
\begin{equation}
Q = \text{Conv}({\tilde{F}_d^{i+}}),
\end{equation}
\begin{equation}
Q'= SN(\text{DWConv}(\text{Conv}({\tilde{F}_d^{i+}})),
\end{equation}
\begin{equation}
G =SN(\Pi^{-1}(Q) W_{{b}}),
\end{equation}
\begin{equation}
Q'' = \mathrm{LN}\!\big(\Pi(Q') \odot G\big),
\end{equation}
where $W_{b}$ denotes the linear projection matrices and $\odot$ denotes element-wise multiplication. The fused representation $Q''$ is modulated by the spike-driven gate $A$ through element-wise multiplication. The result is added to a residual connection with $Q''$ and is projected linearly using $W_{c}$ to obtain $Q'''$,
\begin{equation}
Q''' = W_{c}\big(SN(Q'') \odot A + Q''\big).
\end{equation}
To preserve the original feature information, $Q'''$ is fused with the input feature $F_d^{i+}$,
\begin{equation}
F_d^{i++} = F_d^{i+} + \Pi^{-1}(Q'''),
\end{equation}
where $W_{c}$ is the output projection matrix. 

Finally, the feature $F_d^{i++}$ is processed by a depthwise convolution to form  ${\tilde{F}_d^{i++}}$ as a residual branch. In parallel, the main path applies LayerNorm and SNN-MLP \cite{10848017} to further refine ${\tilde{F}_d^{i++}}$. The final output of the block $\widehat{F}^i_{d}$ is obtained by adding the main path output to the residual branch. It can be formulated as
\begin{equation}
{\tilde{F}_d^{i++}} =\text{DWConv}(F_d^{i++}), 
\end{equation}
\begin{equation}
\widehat{F}^i_{d} = {\tilde{F}_d^{i++}} + \mathrm{MlpSNN}(\mathrm{LN}(\Pi({\tilde{F}_d^{i++}}))).
\end{equation}

\remark{Unlike the conventional attention blocks \cite{vaswani2017attention,dosovitskiy2020image}, 
the SSA block incorporates two spike-driven gates $A$ and $G$ to achieve selective enhancement under sparse conditions. Gate $G$ is applied via element-wise multiplication with locally processed features to dynamically modulate within local region, whereas gate $A$ operates over the global region. To compensate for potential gradient vanishing and information loss caused by gating, multi-level residual connections provide compensation to ensure stable gradient propagation and feature integrity.}

After feature refinement within the SSA block, drone-view and satellite-view features are enhanced to $\widehat{F}_{d}\in \mathbb{R}^{B \times C \times L}$ and $ \widehat{F}_{s}\in \mathbb{R}^{B \times C \times L}$. As illustrated in Fig.~\ref{fig3}, some of $L$ sub-features are modulatedly selected and  serve as emphasized representative cues, denoted as $\widehat{F}^i_{d_1}, \widehat{F}^i_{d_2}, \ldots \widehat{F}^i_{d_n}$, $n\leq L$, and $\widehat{F}^j_{s_1}, \widehat{F}^j_{s_2}, \ldots \widehat{F}^j_{s_m}$, $m\leq L$, respectively. Therefore, the SSA block realizes the selective enhancement of the sub-features. To further reduce the distribution discrepancy of different imaging perspectives,  the features are optimized using cosine embedding loss. This can be described as
\begin{equation}
\mathcal{L}_{\mathrm{1}} = \frac{1}{N}\sum_{k=1}^{N} \Big( 1 - 
\cos\!\big(\widehat{F}^i_{d},\, \widehat{F}^j_{s}\big)\Big),
\end{equation}
where $\cos(\cdot,\cdot)$ denotes cosine similarity and $N$ is the number of paired samples.

\subsection{Spike-Driven Hybrid state space (SHS) Block}
Although the SSA module achieves selective feature enhancement, the resulting cues are still insufficient for DVGL, where geographically distinct locations often exhibit highly similar visual styles. Only the incorporation of long-range dependencies can the model capture the subtle structural differences necessary for reliable cross-view alignment. Recently, state space modules \cite{liu2024vmamba,lee2025efficientvim} have exhibited a strong capability to capture long-range dependencies with only linear complexity. Inspired by \cite{lee2025efficientvim}, the design of state space module is extended into the spiking domain and develops a spike-driven hybrid state space (SHS) block in this work. In the following, the processing of drone-view features is still taken as an example, and the satellite-view features can be obtained in the same way. 

The feature $F^i_d \in \mathbb{R}^{B \times C \times L}$ is reshaped into a 2D feature $F_d^{\eta i} \in \mathbb{R}^{B \times C \times H \times W}$ as input into the SHS block for the subsequent process. As illustrated in Fig.~\ref{fig4}, the SHS block is constructed from three principal components, including two depthwise convolutions, a hybrid state space module (HSM-SSD) \cite{lee2025efficientvim}, and a spike-driven feed-forward network (SFFN). 

A spike-driven depthwise convolution is applied to extract local spatial information with residual connections,
\begin{equation}
F_d^{\tilde{\eta i}} = F_d^{\eta i} + \text{DWConv}(SN(F_d^{\eta i})).
\end{equation}
The feature $F_d^{\tilde{\eta i}} \in \mathbb{R}^{B \times C \times H \times W}$ is reshaped into a 1D sequence $ \mathbb{R}^{B \times C \times L}$ through a flattening operation $Fla(\cdot)$, and subsequently, a LayerNorm is applied to stabilize the feature distribution. Then the HSM-SSD, denoted as $\mathrm{HSM\mbox{-}SSD}(\cdot)$, operates along $L$ to efficiently capture long-range dependencies with linear complexity. After the state space module, the 1D sequence is reshaped by $Res(\cdot, H, W)$ back to a 2D feature $\tilde{F_d^{\eta i+}}\in \mathbb{R}^{B \times C \times H \times W}$. This process can restore the spatial dimensions of the feature that are required for subsequent convolutional operations,
\begin{equation}
\tilde{F_d^{\eta i+}} = Res(\mathrm{HSM\mbox{-}SSD}(\mathrm{LN}(Fla(\tilde{F_d^{\eta i}}))), H,W).
\end{equation}
A secondary spike-driven depthwise convolution is integrated into a residual architecture to enhance local spatial representations and mitigate the loss of fine-grained information in the context of sparse activations. It can be formulated as
\begin{equation}
\tilde{F_d^{\eta i++}} = \tilde{F_d^{\eta i+}} + \text{DWConv}(SN(\tilde{F_d^{\eta i+}})).
\end{equation}
Then the spike-driven feed-forward network $SFFN(\cdot)$ refines channel-wise representations. It is implemented as two $1\times1$ spike-driven convolutions with Batch Normalization,
\begin{equation}
\widehat{F}_d^{\eta i} = \tilde{F_d^{\eta i++}} + {SFFN}(\tilde{F_d^{\eta i++}}).
\end{equation}
\remark{In the SHS block, features are alternately reshaped between sequence and spatial formats. We first reshape the feature from $\mathbb{R}^{B \times C \times L}$ into $\mathbb{R}^{B \times C \times H \times W}$, upon which convolutional operations are performed to inject local spatial priors before entering HSM-SSD module. The feature is then reshaped back to $\mathbb{R}^{B \times C \times L}$, allowing HSM-SSD to efficiently capture long-range dependencies along $L$. Subsequently, the feature is restored to $\mathbb{R}^{B \times C \times H \times W}$ to further learn spatial representations for subsequent processing. In this way, the block jointly captures long-range dependencies and local spatial representations.}

After processing by the SHS block, the drone-view features $\widehat{F}_{d}^{\eta i}$ and satellite-view features $\widehat{F}_{s}^{\eta j}$ are obtained to encode long-range dependencies. As illustrated in Fig.~\ref{fig3}, some of $H\times W$ sub-features within $\widehat{F}_{d}^{\eta i}$, exemplified by $\widehat{F}_{d_1}^{\eta i}$ and $\widehat{F}_{d_2}^{\eta i}$, are encouraged to establish semantic correspondences with their corresponding features 
$\widehat{F}_{s_1}^{\eta i}$ and $\widehat{F}_{s_2}^{\eta j}$. Here $\widehat{F}_{s_1}^{\eta i}$ and $\widehat{F}_{s_2}^{\eta j}$ are the sub-features within $\widehat{F}_{s}^{\eta j}$. By learning these cross-view correspondences, this block enables the learning of consistent semantic relations that transcend mere appearance similarity. This significantly improves performance in accurately identifying location-specific features under view variations. Then both $\widehat{F}_d^{\eta i}$ and $\widehat{F}_s^{\eta j}$ are fed into spike firing activation and subsequently projected by a classification layer. After that, a cross-entropy loss~\cite{xia2024enhancing} is applied to obtain the loss $\mathcal{L}_2$. This supervision enforces discriminative learning of cross-view features.

\subsection{Hierarchical Re-ranking Alignment Learning (HRAL) Strategy}
The two auxiliary blocks inevitably lead to an additional computational cost. Therefore, only the backbone is employed during the inference stage. To achieve  lightweight inference without declining discriminability, an HRAL strategy is proposed, which refines the backbone directly during the training stage. The HRAL strategy combines an improved re-ranking mechanism with contrastive learning. The re-ranking mechanism refines neighborhood structures to provide stronger supervision and improve the consistency between current and historical batches. Meanwhile, contrastive learning drives the backbone features to align with neighborhood representations on different scales. As shown in Fig. \ref{fig3}, the process can be presented as follows.

\textbf{Improved Re-ranking Mechanism.} Let the current batch contain query (drone-view) features $q_f\!\in\!\mathbb{R}^{B_q\times O}$ and gallery (satellite-view) features $g_f\!\in\!\mathbb{R}^{B_g\times O}$, $O= C\times H \times W$. Their are concatenated along the batch dimension to obtain the matrices $F=[q_f;g_f]\in\mathbb{R}^{N\times O}$, $N=B_q\!+\!B_g$, which enables the distance and neighborhood structures between the query and gallery to be computed and optimized in a unified manner. The pairwise distance between $i$-th row and $j$-th row of $F$ is determined,
\begin{equation}
D_{ij} \;=\; \|F_i\|_2^2 + \|F_j\|_2^2 - 2\,F_i^\top F_j,  D\in\mathbb{R}^{N\times N}.
\end{equation}
After performing column-wise normalization on $D_{ij}$ to obtain $\tilde D$, a top-$k$ neighbor list $\mathrm{Rank}(i)$ is generated for each index $i$. Based on this list, the $k$-reciprocal set is constructed as
\begin{equation}\label{eq25}
\mathcal{R}(i) \;=\; \{\,j\in \mathrm{Rank}(i)\;|\; i\in \mathrm{Rank}(j)\,\},
\end{equation}
which is further expanded by incorporating candidates whose reciprocal sets exhibit sufficient overlap with $\mathcal{R}(i)$, and the overlap ratio should be greater than $\tfrac{2}{3}$. This produces robust local neighborhoods that are mutually consistent rather than one-sided. According to $\mathcal{R}(i)$, a row-stochastic affinity $V\!\in\!\mathbb{R}^{N\times N}$ with Gaussian weights $\varepsilon$ is built on normalized distances,
\begin{equation}
V_{ij} \;=\;
\begin{cases}
\displaystyle \frac{\exp(-\tilde D_{ij})}{\sum_{v\in \mathcal{R}(i)} \exp(-\tilde D_{iv}) + \varepsilon}, & j\in \mathcal{R}(i),\\
0, & \text{otherwise}.
\end{cases}
\end{equation}
To further stabilize the weights, a light query-expansion smoothing is applied, in which the rows of $V$ over the initial top-$k$ neighbors are averaged to replace the original $V$ with its smoothed version. Finally, refined features are obtained using residual diffusion and L2-normalization $\mathrm{Norm}_{\ell_2}(\cdot)$,
\begin{equation}\label{eq27}
\hat F =\mathrm{Norm}_{\ell_2}( \alpha\,F + (1-\alpha)\,V F),
\end{equation}
where $\alpha\!\in\![0,1]$ is set to 0.7. The refined query and gallery features, denoted as $\hat q_f$ and $\hat g_f$, are obtained by splitting $\hat F$ according to $B_q$ and $B_g$.

\textbf{Hierarchical Alignment.} Relying solely on the current batch limits re-ranking to a narrow and potentially biased feature distribution. To address this, a memory-based mechanism is proposed to store query and gallery features from historical batches. During the training stage, the improved re-ranking mechanism is applied to both the current batch $(q_f,g_f)$ and historical batches $(q_f^{\mathcal Q}, g_f^{\mathcal G})$. As a result,  $(\hat q_f,\hat g_f)$ is obtained for the current batch and $(\hat q_f^{\mathcal Q}, \hat g_f^{\mathcal G})$ is obtained for historical samples. By integrating neighborhood relationships between current and historical batches, the re-ranking process is extended to a broader and more representative feature space.

\textbf{Contrastive Learning.} Although the improved re-ranking mechanism is unable to participate in the training stage, its supervisory effect can be represented through explicit alignment losses that compare the original features with their refined features. At the batch level, consistency is aligned between the features within the current batch and their re-ranked counterparts. Specifically, each feature must remain close to its refined feature based on cosine similarity and Kullback–Leibler (KL) divergence. The current batch loss $\mathcal{L}_{\mathrm{current}}$ is formulated as
\begin{equation}
\mathcal{L}^{q}_{\cos} = 1 - \cos(q_f, \hat q_f),
\end{equation}
\begin{equation}
\mathcal{L}^{g}_{\cos} = 1 - \cos(g_f, \hat g_f),
\end{equation}
\begin{equation}
\mathcal{L}^{q}_{\mathrm{KL}}= \mathrm{KL}\!\big(p(q_f \hat q_f^\top) \;\|\; p(\hat q_f \hat q_f^\top)\big),
\end{equation}
\begin{equation}
\mathcal{L}^{g}_{\mathrm{KL}} = \mathrm{KL}\!\big(p(g_f \hat g_f^\top) \;\|\; p(\hat g_f \hat g_f^\top)\big),
\end{equation}
\begin{equation}
\mathcal{L}_{\mathrm{current}}=\mathcal{L}^{q}_{\cos} + \mathcal{L}^{q}_{\mathrm{KL}} + \mathcal{L}^{g}_{\cos} + \mathcal{L}^{g}_{\mathrm{KL}},
\end{equation}
where $p(\cdot)$ denotes the normalized similarity distribution calculated using a softmax function. Then, the batch level is extended from the current batch to historical batches, and the historical batch loss $\mathcal{L}_{\mathrm{historical}}$ can be obtained in the same manner. $\mathcal{L}_{\mathrm{historical}}$ as a complement to the current batch, further enhances alignment across long-term neighborhood structures to maintain cross-batch consistency. Additionally, to directly encourage cross-view feature alignment, InfoNCE loss~\cite{oord2018representation} is adopted to align $q_f$ and $g_f$. The HRAL strategy combines these three losses as the optimization objective,
\begin{equation}
\mathcal{L}_{\mathrm{3}} =\mathcal{L}_{\mathrm{current}} + \mathcal{L}_{\mathrm{historical}} +\mathcal{L}_{\mathrm{InfoNCE}}.
\end{equation}

The proposed SNN framework is trained in an end-to-end manner by jointly optimizing $\mathcal{L}_1$, $\mathcal{L}_2$ together with $\mathcal{L}_{\mathrm{3}}$. The total objective can be expressed as
\begin{equation}\label{eq34}
\mathcal{L}_{\mathrm{total}} = \lambda_1\mathcal{L}_1 + \lambda_2\mathcal{L}_2 +\mathcal{L}_3,
\end{equation}
where $\lambda_1$ and $\lambda_2$ are weight coefficients. 

\remark{Unlike conventional re-ranking \cite{zhong2017re} that is only applied as inference heuristic, the proposed HRAL strategy recasts re-ranking as supervised strategy during training. At each iteration, the features extracted by the backbone are trained to align with their refined features through re-ranking. This turns re-ranking from a passive post-processing step into an active supervisory signal. Spurious neighbors are pruned via $k$-reciprocal consistency, reliable context is propagated through Gaussian weighting and query-expansion smoothing, and refinement is conservatively fusion with the parameter $\alpha$. Consequently, HRAL enhances cross-view alignment while ensuring stable optimization.}

\subsection{Theoretical Energy Consumption Analysis}\label{energy}
Energy efficiency is one of the most critical advantages of SNNs, due to their event-driven paradigm in which computation is activated only upon spike firing. Compared with ANNs, this naturally leads to sparse computation and significantly reduces energy cost. In the proposed SpikeViMFormer, the energy efficiency would be verified by energy consumption analysis. 

Following prior work \cite{molchanov2016pruning}, the computational burden in ANNs is typically measured in terms of floating-point operations (FLOPs). For a convolution layer and an MLP layer, the FLOPs can be expressed as
\begin{equation}
FL_{\mathrm{Conv}} = (k_n)^2 \cdot h_n \cdot w_n \cdot c_{n-1} \cdot c_n,
\end{equation}
\begin{equation}
FL_{\mathrm{MLP}} = i_m \cdot o_m,
\end{equation}
where $k_n$ is the kernel size and $(h_n,w_n)$ are the output feature dimensions. $c_{n-1}$ and $c_n$ denote the input and output channel numbers. $i_m$ and $o_m$ denote the input and output dimensions of the $m$-th MLP layer.

In general SNNs, these FLOPs are converted into sparse computation. The energy consumption is proportional to the number of simulation timesteps $T$ and the spike firing rate $R_f$, which is the proportion of non-zero elements in the spike tensor. The energy consumption can be approximated as
\begin{equation}\label{eq37}
E_{\mathrm{SNN}} \approx E_{AC} \times T \times R_f \times FL,
\end{equation}
where $E_{AC}$ denotes the energy of a single AC operation, and $FL$ is the corresponding FLOPs of the ANN counterpart. In comparison, ANNs require the energy $E_{MAC}$ of a single MAC operation. 
\begin{equation}
E_{\mathrm{ANN}} \approx E_{MAC} \times FL.
\end{equation}

Therefore, SNNs are theoretically more energy-efficient whenever
\begin{equation}
E_{AC} \times T \times R_f < E_{MAC}.
\end{equation}

Based on previous hardware measurements \cite{yin2021accurate,yao2023attention}, the energy cost of $E_{MAC}$ is approximately $4.6$pJ, whereas that of $E_{AC}$ is only about $0.9$pJ under 45nm CMOS implementations. Given that the spike firing rate $R$ is typically much lower than $1$ due to sparse activations, $E_{AC} \times T \times R_f < E_{MAC}$ generally holds in practice to ensure that SNNs achieve substantially lower energy consumption than their ANN counterparts.

\section{Experimental results}\label{results}
\subsection{Experimental Datasets and Evaluation Metrics}
\textbf{Experimental Datasets}. To evaluate the proposed SpikeViMFormer, we perform experiments on two benchmark datasets, including University-1652 \cite{zheng2020university} and SUES-200 \cite{zhu2023sues}. University-1652 is a large-scale dataset comprising images of 1,652 buildings collected from 72 universities around the world. It involves drone-view, satellite-view, and ground-view images. The training partition contains 701 buildings from 33 universities, while the testing partition includes 951 buildings from 39 universities without geographical overlap. SUES-200 emphasizes aerial imagery at diverse flight altitudes. It covers 200 locations, with 120 used for training and 80 for testing. Each location provides a single satellite image and several aerial images taken from four altitude levels, which are intended to evaluate cross-view retrieval under different altitude conditions.

\textbf{Evaluation Metrics}. In this work, we report the Recall@K(R@k) and the mean average precision (mAP) as the primary evaluation metrics. R@k quantifies the percentage of queries for which the ground-truth reference image is found within the highest $k$ retrieved candidates, while mAP accounts for the ranking quality by calculating the average precision of each query and then averaging over the entire test set. Furthermore, all methods are evaluated in terms of their energy consumption, as discussed in Section \ref{energy}, and their parameter count. Additionally, the timesteps of all directly trained SNNs are represented in the format $T \times D$, where $T$ denotes the number of timesteps and $D$ has been defined in Eq.(\ref{eq5}). Notably, in the experimental tables, the columns ``Param'', ``Power'', and ``Step'' correspond to ``Parameter count'', ``Energy consumption per inference'', and ``Timesteps'', respectively.

\subsection{Implementation Details}
All input images are resized to $384 \times 384$. SpikeViMFormer is trained for a total of 5 epochs with a batch size of 64. It is optimized using the AdamW optimizer with an initial learning rate of $1 \times 10^{-4}$. In the HRAL strategy, the hyper-parameter $k$ used in the  top-$k$ neighbor list is set to 15. For the loss function in Eq.(\ref{eq34}), the coefficients $\lambda_{1}$ and $\lambda_{2}$ are set to 0.6 and 0.54, respectively. All experiments are implemented using the PyTorch framework and are conducted on an Ubuntu 22.04 platform equipped with eight NVIDIA RTX 4090 GPUs.

\begin{table}[t]
\caption{Comparison of SpikeViMFormer with Advanced ANNs and State-of-the-art SNNs on University-1652.}
\centering
\large
\resizebox{\columnwidth}{!}{ 
\setlength{\tabcolsep}{3pt}
\begin{tabular}{ccccccccc}\hline
\multicolumn{9}{c}{University-1652}  \\ \hline
\multirow{2}{*}{Framework} &\multirow{2}{*}{Method} & \multirow{2}{*}{\makecell[c]{Param \\(M)}} & \multirow{2}{*}{\makecell[c]{Power \\(mJ)}} & \multirow{2}{*}{\makecell[c]{Step}} & \multicolumn{2}{c}{Drone$\rightarrow$Satellite} & \multicolumn{2}{c}{Satellite$\rightarrow$Drone} \\ \cline{6-9} 
&  &&& & R@1 & AP   & R@1 & AP\\ \hline
\multirow{5}{*}{\text{ANN}} 
  &LPN\cite{wang2021each}           & 62.39     & 169.19 &1  & 75.93       & 79.14      & 86.45     & 74.49 \\
&TransFG\cite{zhao2024transfg}       & -  & -  & 1 & 84.01       & 86.31      & 90.16     & 84.61 \\
& FSRA\cite{dai2021transformer}	& 52.01	& 113.11	& 1	& 84.51	& 86.71	& 84.51	& 86.71 \\
&  IFSs\cite{ge2024multibranch}       & -  & -     & 1   & 86.06       & 88.08      & 91.44     & 85.73 \\
&SDPL\cite{chen2024sdpl}           &42.56  & 320.67  &1 & 90.16       & 91.64      & 93.58     & 89.45 \\
&CCR\cite{du2024ccr}  &156.57 & 738.10  & 1 &91.31       &92.70     & 94.58 &90.60 \\\hline
\multirow{6}{*}{\text{SNN}} & Meta-SpikeFormer\cite{yao2024spikedriven}   & 55.35 & 308.28 & $1\times 4$ & 78.94 & 82.07 &88.59 & 79.12 \\\cline{2-9}
& {E-SpikeFormer-1\cite{10848017}}

 & 171.32 & 706.31 & $1\times 8$ & 81.50 & 84.54 &91.01 & 80.99 \\
&{E-SpikeFormer-2\cite{10848017}} & 171.32 & 700.31 & $1\times 8$ & 82.49 & 85.48 &90.58 & 80.52 \\\cline{2-9} 
&{SpikeViMFormer-T} &9.78	&32.53	&$1\times4$	&86.10	&88.32	&91.58	&85.40 \\
&SpikeViMFormer-S &18.63	&53.88 	&$1\times4$	&\textbf{88.03}	&\textbf{89.98}	&\textbf{92.72}	&\textbf{87.10}\\ \hline
\end{tabular}}
\label{tab:university}
\footnote{1}{\tiny E-SpikeFormer-1 and E-SpikeFormer-2 denote the two configurations of E-SpikeFormer provided in \cite{10848017}.}
\end{table}

\begin{table*}[t]
\centering
\tiny
\caption{Comparison of SpikeViMFormer with Advanced ANNs and State-of-the-art SNNs on SUES-200.}
  \setlength{\tabcolsep}{3pt}
\resizebox{\textwidth}{!}{%
\begin{tabular}{ccccccccccccc|cccccccc}
\hline
&\multicolumn{19}{c}{SUES-200} \\ \hline
\multirow{3}{*}{Framework} & \multirow{3}{*}{Method} & \multirow{3}{*}{\makecell[c]{Param\\(M)}} & \multirow{3}{*}{\makecell[c]{Power \\(mJ)}} & \multirow{3}{*}{\makecell[c]{Step}} & \multicolumn{8}{c|}{Drone$\rightarrow$Satellite} & \multicolumn{8}{c}{Satellite$\rightarrow$Drone} \\ \cline{6-21} 
                && &        &                                & \multicolumn{2}{c}{150m} & \multicolumn{2}{c}{200m} & \multicolumn{2}{c}{250m} & \multicolumn{2}{c|}{300m} & \multicolumn{2}{c}{150m} & \multicolumn{2}{c}{200m} & \multicolumn{2}{c}{250m} & \multicolumn{2}{c}{300m} \\ \cline{6-21} 
            && &            &                                & R@1 & AP   & R@1 & AP   & R@1 & AP   & R@1 & AP   & R@1 & AP   & R@1 & AP   & R@1 & AP   & R@1 & AP \\ \hline
\multirow{6}{*}{ANN}&LPN\cite{wang2021each} &62.39 &169.19  &1 &61.58 &67.23 &70.85 &75.96 &80.38 &83.80 &81.47 &84.53 &83.75 &66.78 &88.75 &75.01 &92.50 &81.34 &92.50 &85.72\\
& FSRA\cite{dai2021transformer} &52.01 &113.11  &1 &68.25	&73.45	&83.00	&85.99	&90.68	&92.27	&91.95	&93.46 &83.75	&76.67	&90.00	&85.34	&93.75	&90.17	&95.00	&92.03\\
& IFSs \cite{ge2024multibranch}    &- &- &1 &77.57 &81.30 &89.50 &91.40 &92.58 &94.21 &97.40 &97.92 &93.75 &79.49 &{97.50} &90.52 &97.50 &96.03 &{100.00} &97.66\\ 
&SDPL\cite{chen2024sdpl}  &42.56 &320.67 &1 &82.95 &85.82 &92.73 &94.07 &96.05 &96.69 &97.83 &98.05 &93.75 &83.75 &96.25 &92.42 &97.50 &95.65 &96.25 &96.17 \\
  &CCR\cite{du2024ccr}  & 156.57  &738.10 &1 & 87.08 & 89.55 & 93.57 & 94.90 & 95.42 & 96.28 & 96.82 & 97.39 & 92.50 & 88.54 & 97.50 & 95.22 & 97.50 & 97.10 & 97.50 & 97.49\\
\hline
\multirow{5}{*}{SNN} & Meta-SpikeFormer\cite{yao2024spikedriven}  &55.35  &303.14 &$ 1\times4$ &76.62 &81.09 &86.37 & 89.25  &88.77 &91.24 &90.27 &92.38 &85.00 &77.42 &91.25 &87.10 &93.75 &89.67 &{93.75} &91.14\\\cline{2-21} 

& \multirow{1}{*}{E-SpikeFormer-1\cite{10848017}}  & 171.32 &716.16 &$ 1\times8 $ & 77.57 & 81.52 & 86.20 & 88.96 & 90.17 & 92.16 & 91.95 & 93.61 & 86.25 & 80.27 & 91.25 & 87.69 & 97.50 & 92.65 & 96.25 & 94.26\\ 
  
  &E-SpikeFormer-2\cite{10848017} &171.32 &710.18 & $ 1\times8 $ &77.55 &81.43 &86.65 &89.20 &90.22 &92.14 &92.00 &93.59 &91.25 &79.65 &{92.50} &87.32 &95.00 &91.41 &{96.25} &92.85\\ \cline{2-21} 
& {SpikeViMFormer-T}  &9.78 &32.69 &$ 1\times4 $ &75.80	&80.13	&84.53	&87.44	&89.75	&91.63	&91.10	&92.70 &88.75	&79.57	&91.25	&87.21	&92.50	&90.53	&92.50	&90.49\\
  &SpikeViMFormer-S & 18.63  &54.33 &$ 1\times4 $ &\textbf{83.48} &\textbf{86.51} &\textbf{89.65} &\textbf{91.74} &\textbf{92.68} &\textbf{94.09} &\textbf{93.90} &\textbf{95.08} &\textbf{92.50} &\textbf{84.54} &\textbf{97.50} &\textbf{91.62} &\textbf{96.25} &\textbf{93.89} &\textbf{96.25} &\textbf{95.10}\\\hline
\end{tabular}%
}
\label{tab:SUES-200}
\end{table*}

\subsection{Comparison with State-of-the-art Methods}
SpikeViMFormer uses two backbone variants from \cite{10848017} with parameter counts of 10M and 19M, respectively. Their classification layers are pruned in this work. Both backbones are optimized using InfoNCE loss to serve as baselines. For clarity, the SpikeViMFormer version with the smaller backbone is denoted as SpikeViMFormer-T, and the SpikeViMFormer version with the larger backbone is denoted as SpikeViMFormer-S, corresponding to baseline-T and baseline-S, respectively.

Table \ref{tab:university} shows the comparisons of SpikeViMFormer with advanced ANNs including LPN \cite{wang2021each}, TransFG \cite{zhao2024transfg}, FSRA \cite{dai2021transformer}, IFSs \cite{ge2024multibranch}, SDPL \cite{chen2024sdpl} and CCR \cite{du2024ccr}, and state-of-the-art SNNs including Meta-SpikeFormer \cite{yao2024spikedriven} and E-SpikeFormer \cite{10848017} on the University-1652 dataset. Additionally, Table \ref{tab:SUES-200} presents the results on the SUES-200 dataset.

\textbf{Comparison with ANNs.}
On the University-1652 dataset, compared with advanced ANNs, the proposed SpikeViMFormer achieves superior performance with two versions. SpikeViMFormer-T requires only 32.53 mJ per inference and outperforms LPN, FSRA, TransFG and IFS in R@1 and AP. It reduces parameter count by approximately 5.3× and energy consumption by 3.5× compared with FSRA. Compared with the more advanced SDPL, SpikeViMFormer-S experiences only 1–2\% accuracy reduction, but achieves substantial reductions in both parameter count and energy consumption. Compared with CCR, SpikeViMFormer-S exhibits a 3\% relative accuracy trade-off, but reduces energy consumption of inference stage by 13.24× and parameter count by 8.4×. On the SUES-200 dataset, the proposed SpikeViMFormer exhibits a similarly compelling performance. Notably, SpikeViMFormer-S surpasses most advanced ANNs, including LPN, FSRA, and IFS. It also maintains highly competitive accuracy with extremely low parameter count and energy consumption compared with SDPL and CCR.

\textbf{Comparison with SNNs.}
Compared with SOTA SNNs including Meta-SpikeFormer and E-SpikeFormer on the University-1652 dataset, SpikeViMFormer-T outperforms them in all metrics. It reduces the parameter count by approximately 5.7× and 18.1×, and the energy consumption of the inference stage by 9.6× and 21.8×, respectively. In particular, compared with SpikeViMFormer-T, SpikeViMFormer-S incurs only a modest increase in parameter count and energy consumption, but its accuracy improves significantly by approximately 6\% in R@1 and 4\% in AP relative to SOTA SNNs, respectively. On the SUES-200 dataset, while SpikeViMFormer-T does not outperform Meta-SpikeFormer and E-SpikeFormer, it maintains a significantly lower parameter count and lower energy consumption. More importantly, SpikeViMFormer-S surpasses both Meta-SpikeFormer and E-SpikeFormer in all 16 evaluation metrics.

\begin{table*}[t]
\centering
\tiny
\caption{Performance of Integrated State-of-the-art SNNs within the Proposed SpikeViMFormer on SUES-200.}
  \setlength{\tabcolsep}{3pt}
\resizebox{\textwidth}{!}{%
\begin{tabular}{ccccccccccccc|cccccccc}
\hline
&\multicolumn{19}{c}{SUES-200} \\ \hline
\multirow{3}{*}{Framework} & \multirow{3}{*}{Method} & \multirow{3}{*}{\makecell[c]{Param\\(M)}} & \multirow{3}{*}{\makecell[c]{Power \\(mJ)}} & \multirow{3}{*}{\makecell[c]{Step}} & \multicolumn{8}{c|}{Drone$\rightarrow$Satellite} & \multicolumn{8}{c}{Satellite$\rightarrow$Drone} \\ \cline{6-21} 
                && &        &                                & \multicolumn{2}{c}{150m} & \multicolumn{2}{c}{200m} & \multicolumn{2}{c}{250m} & \multicolumn{2}{c|}{300m} & \multicolumn{2}{c}{150m} & \multicolumn{2}{c}{200m} & \multicolumn{2}{c}{250m} & \multicolumn{2}{c}{300m} \\ \cline{6-21} 
            && &            &                                & R@1 & AP   & R@1 & AP   & R@1 & AP   & R@1 & AP   & R@1 & AP   & R@1 & AP   & R@1 & AP   & R@1 & AP \\ \hline
\multirow{6}{*}{SNN}&{Meta-SpikeFormer\cite{yao2024spikedriven}} &55.35  &303.14 &$ 1\times4$ &76.62 &81.09 &86.37 & 89.25  &88.77 &91.24 &90.27 &92.38 &85.00 &77.42 &91.25 &87.10 &93.75 &89.67 &{93.75} &91.14 \\
&Meta-SpikeFormer$^{\scalebox{0.5}{$\dagger$}}$ &55.35  &303.20 &$ 1\times4 $ &\textbf{78.18} &\textbf{82.30} &\textbf{88.18} &\textbf{90.49} &\textbf{92.35} &\textbf{93.72} &\textbf{94.23} &\textbf{95.11} &\textbf{85.00} &\textbf{78.01} &\textbf{93.75} &\textbf{87.70} &92.50 &\textbf{90.31}	&92.50 &\textbf{91.70}\\ \cline{2-21}
& \multirow{1}{*}{E-SpikeFormer-1\cite{10848017}}& 171.32  &716.16 &$ 1\times8 $ & 77.57 & 81.52 & 86.20 & 88.96 & 90.17 & 92.16 & 91.95 & 93.61 & 86.25 & 80.27 & 91.25 & 87.69 & 97.50 & 92.65 & 96.25 & 94.26\\ 
&\multirow{1}{*}{E-SpikeFormer-1$^{\scalebox{0.5}{$\dagger$}}$ } &171.32 &724.05 &$ 1\times8 $ &\textbf{78.32} &\textbf{81.83} &\textbf{86.83} &\textbf{89.10} &\textbf{90.32} &91.70 &\textbf{92.01} &\textbf{93.72} &\textbf{88.75} &\textbf{80.30} &\textbf{97.50} &\textbf{89.19} &96.25 &\textbf{92.68} &\textbf{96.25} &93.32\\\cline{2-21}
  &E-SpikeFormer-2\cite{10848017} &171.32 &710.18 & $ 1\times8 $ &77.55 &81.43 &86.65 &89.20 &90.22 &92.14 &92.00 &93.59 &91.25 &79.65 &{92.50} &87.32 &95.00 &91.41 &{96.25} &92.85\\ 
&E-SpikeFormer-2$^{\scalebox{0.5}{$\dagger$}}$   &171.32 &718.90 &$ 1\times8 $ &\textbf{79.08} &\textbf{82.89} &\textbf{89.13} &\textbf{91.28} &\textbf{93.35} &\textbf{94.77} &\textbf{93.70} &\textbf{95.07} &\textbf{92.50} &\textbf{81.45} &\textbf{95.00} &\textbf{90.40} &\textbf{96.25} &\textbf{93.84} &\textbf{96.25} &\textbf{94.20}\\\hline

\end{tabular}%
}
\label{tab:SUES-200improve}
\end{table*}

\subsection{Generalizability Assessment}
To assess the generalizability, the backbone of SpikeViMFormer is replaced by other SOTA SNNs for comparison. Meta-SpikeFormer$^{\scalebox{0.5}{$\dagger$}}$  and E-SpikeFormer$^{\scalebox{0.5}{$\dagger$}}$  denote variants of SpikeViMFormer where the backbones are replaced with Meta-SpikeFormer and E-SpikeFormer, respectively. The evaluation results for both datasets are shown in Tables \ref{tab:SUES-200improve} and \ref{tab:withours}.

\textbf{Generalizability on SUES-200}. As shown in Table \ref{tab:SUES-200improve}, the Meta-SpikeFormer$^{\scalebox{0.5}{$\dagger$}}$ maintains nearly constant energy consumption but achieves improvements in 14 out of 16 evaluation metrics on the SUES-200 dataset. For the E-SpikeFormer-1$^{\scalebox{0.5}{$\dagger$}}$ and E-SpikeFormer-2$^{\scalebox{0.5}{$\dagger$}}$, a marginal increase in energy consumption is observed. Nevertheless, the E-SpikeFormer-1$^{\scalebox{0.5}{$\dagger$}}$ improves 13 metrics, while the E-SpikeFormer-2$^{\scalebox{0.5}{$\dagger$}}$ exhibits improvements in all evaluation metrics.

\begin{table}[t]
\caption{Performance of Integrated State-of-the-art SNNs within the Proposed SpikeViMFormer on University-1652.}
\centering
\large
\resizebox{\columnwidth}{!}{ 
\setlength{\tabcolsep}{3pt}
\begin{tabular}{ccccccccc}\hline
\multicolumn{9}{c}{University-1652}  \\ \hline
\multirow{2}{*}{Framework} &\multirow{2}{*}{Method} & \multirow{2}{*}{\makecell[c]{Param\\ (M)}} & \multirow{2}{*}{\makecell[c]{Power \\(mJ)}} & \multirow{2}{*}{\makecell[c]{Step}} & \multicolumn{2}{c}{Drone$\rightarrow$Satellite} & \multicolumn{2}{c}{Satellite$\rightarrow$Drone} \\ \cline{6-9} 
&  &&& & R@1 & AP   & R@1 & AP\\ \hline
\multirow{6}{*}{\text{SNN}} & Meta-SpikeFormer\cite{yao2024spikedriven}   & 55.35 &308.28 & $1\times 4$ & 78.94 & 82.07 &88.59 & 79.12 \\
&  Meta-SpikeFormer$^{\scalebox{0.5}{$\dagger$}}$   & 55.35 & 303.04 & $1\times 4$ & \textbf{89.10} &\textbf{90.74} &\textbf{92.72} & \textbf{88.32} \\\cline{2-9}
& \multirow{1}{*}{E-SpikeFormer-1\cite{10848017}}& 171.32 & 706.31& $1\times 8$ & 81.50 & 84.54 &91.01 & 80.99 \\
&\multirow{1}{*}{E-SpikeFormer-1$^{\scalebox{0.5}{$\dagger$}}$ } &171.32	&721.85	&$1\times8$	& \textbf{82.60}	&\textbf{85.54}	&90.16	&\textbf{81.34} \\\cline{2-9}
&\multirow{1}{*}{E-SpikeFormer-2\cite{10848017}} & 171.32 & 700.31& $1\times 8$ &{82.49} & {85.48} &90.58 & 80.52 \\
&\multirow{1}{*}{E-SpikeFormer-2$^{\scalebox{0.5}{$\dagger$}}$ }&171.32	&720.00	&$1\times8$	&\textbf{83.61}	&\textbf{86.41}	&\textbf{91.73}	&\textbf{82.22}\\ \hline
\end{tabular}}
\label{tab:withours}
\end{table}

\textbf{Generalizability on University-1652}. 
As shown in Table \ref{tab:withours}, the Meta-SpikeFormer$^{\scalebox{0.5}{$\dagger$}}$ achieves a remarkable improvement. Notably, the R@1 and AP metrics for the drone$\rightarrow$satellite scenario increase by 10.16\% and 8.67\%, respectively, while for the satellite$\rightarrow$drone scenario, they increase by 4.13\% and 9.2\%, respectively. Furthermore, energy consumption is reduced by 5.24 mJ. Additionally, although E-SpikeFormer-1$^{\scalebox{0.5}{$\dagger$}}$ and E-SpikeFormer-2$^{\scalebox{0.5}{$\dagger$}}$ incur a minor rise in energy consumption, R@1 and AP have improved. 

These results demonstrate that the SpikeViMFormer consistently improves performance with different backbones.

\begin{table}[t]
\caption{Comparisons between baseline and SpikeViMFormer on University-1652.}
\centering
\large
\resizebox{\columnwidth}{!}{ 
\setlength{\tabcolsep}{3pt}
\begin{tabular}{ccccccccc}\hline
\multicolumn{9}{c}{University-1652}  \\ \hline
\multirow{2}{*}{Framework} &\multirow{2}{*}{Method} & \multirow{2}{*}{\makecell[c]{Param\\ (M)}} & \multirow{2}{*}{\makecell[c]{Power \\(mJ)}} & \multirow{2}{*}{\makecell[c]{Step}} & \multicolumn{2}{c}{Drone$\rightarrow$Satellite} & \multicolumn{2}{c}{Satellite$\rightarrow$Drone} \\ \cline{6-9} 
&  &&& & R@1 & AP   & R@1 & AP\\ \hline
\multirow{4}{*}{\text{SNN}} &{Baseline-T}   & 9.78 &32.58 & $1\times 4$ &76.59	&80.39	&89.01	&76.25 \\
&Baseline-S& 18.63 &55.69 & $1\times 4$ & 78.65 & 82.12 &87.73 & 78.64 \\\cline{2-9}
& {SpikeViMFormer-T}
& 9.78 &32.53 & $1\times 4$ &  \textbf{86.10}  & \textbf{88.32} & \textbf{91.58}  &\textbf{85.40}\\
 &SpikeViMFormer-S& 18.63 &53.88 & $1\times 4$ & \textbf{88.03} & \textbf{89.98} &\textbf{92.72} & \textbf{87.10} \\ \hline
\end{tabular}}
\label{tab:baseline}
\end{table}

\subsection{Ablation Studies}
In this subsection, we conduct a comprehensive set of ablation studies. The performance of SpikeViMFormer is first compared with the baseline to validate its overall effectiveness. The corresponding results are reported in Tables \ref{tab:baseline} and \ref{tab:SUES-200baseline}.

\textbf{Comparison with Baseline on University-1652}. As shown in Table \ref{tab:baseline}. The Baseline-T, with 9.78M parameter count and 32.58mJ, achieves R@1 of 76.59\% and AP of 80.39\% for drone$\rightarrow$satellite scenario, and R@1 of 89.01\% and AP of 76.25\% for satellite$\rightarrow$drone scenario.  SpikeViMFormer-T outperforms it with R@1 of 86.10\% (+9.51\%) and AP of 88.32\% (+7.93\%) for drone$\rightarrow$satellite scenario, and R@1 of 91.58\% (+2.57\%) and AP of 85.40\% (+9.15\%) for satellite$\rightarrow$drone scenario. Compared with the Baseline-S, SpikeViMFormer-S further improves R@1 by 88.03\% (+9.38\%) and AP by 89.98\% (+7.86\%) for drone$\rightarrow$satellite scenario, and R@1 by 92.72\% (+4.99\%) and AP by 87.10\% (+8.46\%) for satellite$\rightarrow$drone scenario. Notably, SpikeViMFormer achieves energy consumption that is on a par with, or even lower than, the baseline, which demonstrates the effectiveness of SpikeViMFormer over the baseline without losing energy efficiency for performance improvement.

\begin{table*}[t]
\centering
\tiny
\caption{Comparisons between baseline and SpikeViMformer on SUES-200.}
  \setlength{\tabcolsep}{3pt}
\resizebox{\textwidth}{!}{%
\begin{tabular}{ccccccccccccc|cccccccc}
\hline
&\multicolumn{19}{c}{SUES-200} \\ \hline
\multirow{3}{*}{Framework} & \multirow{3}{*}{Method} & \multirow{3}{*}{\makecell[c]{Param\\(M)}} & \multirow{3}{*}{\makecell[c]{Power \\(mJ)}} & \multirow{3}{*}{\makecell[c]{Step}} & \multicolumn{8}{c|}{Drone$\rightarrow$Satellite} & \multicolumn{8}{c}{Satellite$\rightarrow$Drone} \\ \cline{6-21} 
                && &        &                                & \multicolumn{2}{c}{150m} & \multicolumn{2}{c}{200m} & \multicolumn{2}{c}{250m} & \multicolumn{2}{c|}{300m} & \multicolumn{2}{c}{150m} & \multicolumn{2}{c}{200m} & \multicolumn{2}{c}{250m} & \multicolumn{2}{c}{300m} \\ \cline{6-21} 
            && &            &                                & R@1 & AP   & R@1 & AP   & R@1 & AP   & R@1 & AP   & R@1 & AP   & R@1 & AP   & R@1 & AP   & R@1 & AP \\ \hline
\multirow{4}{*}{SNN}&{Baseline-T} &9.78 &32.49 &$ 1\times4 $ &74.55 &78.93 &83.82 &86.85 &88.32 &90.60 &90.05 &91.92 &85.00 &74.30 &95.00 &84.60 &96.25 &90.52 &95.00 &91.83\\
&Baseline-S&18.63 &55.27 &$ 1\times4 $ &74.75 &79.03 &87.25 &89.50 &91.32&92.91 &92.02 &93.54 &95.00 &78.95 &98.75 &89.44 &97.50 &91.62 &95.00 &91.92\\ 
 \cline{2-21}
&{SpikeViMFormer-T}  &9.78 &32.69 &$ 1\times4 $ &\textbf{75.80}	&\textbf{80.13}	&\textbf{84.53}	&\textbf{87.44}	&\textbf{89.75}	&\textbf{91.63}	&\textbf{91.10}	&\textbf{92.70} &\textbf{88.75}	&\textbf{79.57}	&\textbf{91.25}	&\textbf{87.21}	&92.50	&\textbf{90.53}	&92.50	&90.49\\
  &{SpikeViMFormer-S} & 18.63  &54.33 &$ 1\times4 $ &\textbf{83.48} &\textbf{86.51} &\textbf{89.65} &\textbf{91.74} &\textbf{92.68} &\textbf{94.09} &\textbf{93.90} &\textbf{95.08} &92.50 &\textbf{84.54} &97.50 &\textbf{91.62} &96.25 &\textbf{93.89} &\textbf{96.25} &\textbf{95.10}\\\hline

\end{tabular}%
}
\label{tab:SUES-200baseline}
\end{table*}

\begin{table}[t]
\centering
\caption{Influence of each component on performance of SpikeViMFormer.}
\tiny
\resizebox{\columnwidth}{!}{ 
\setlength{\tabcolsep}{3pt}
\begin{tabular}{lcccccccc}\hline
\multicolumn{8}{c}{University-1652}  \\ \hline
\multirow{2}{*}{Method} & \multirow{2}{*}{SSA} & \multirow{2}{*}{SHS} & \multirow{2}{*}{HRAL} & \multicolumn{2}{c}{Drone$\rightarrow$Satellite} & \multicolumn{2}{c}{Satellite$\rightarrow$Drone} \\ \cline{5-8}
&&&& R@1 & AP & R@1 & AP \\ \hline
\multirow{5}{*}{{SpikeViMFormer-T}} &  &  &  &76.59	&80.39	&89.01	&76.25 \\
 & $\checkmark$ &  &  &{77.85} & {81.31} & {89.87} & {78.44} \\
  &  & $\checkmark$ & & {82.13} & {85.01} &{90.72}  & {81.72} \\
  & $\checkmark$ &$\checkmark$  &  &{84.31} &{86.89} &\textbf{92.43} & {84.39} \\
 & $\checkmark$ & $\checkmark$ &$\checkmark$  &  \textbf{86.10}  & \textbf{88.32} & 91.58  &\textbf{85.40}  \\ \hline
 \multirow{5}{*}{{SpikeViMFormer-S}} &  &  & & 78.65 & 82.12 &87.73 & 78.64 \\
 & $\checkmark$ &  &  & {82.44} & {85.29} & {90.44} & {81.58} \\
  &  &$\checkmark$  &  & {85.87} & {88.23} & {92.43} & {85.17} \\
 & $\checkmark$ & $\checkmark$ & &87.21       &89.35 &\textbf{93.29}     &{86.09}\\  & $\checkmark$ & $\checkmark$ &$\checkmark$  & \textbf{88.03} & \textbf{89.98} &{92.72} & \textbf{87.10} \\ \hline
\end{tabular}}
\label{tab:university-ab}
\end{table}

\textbf{Comparison with Baseline on SUES-200}. As shown in Table \ref{tab:SUES-200baseline}, in the 16 evaluation metrics on the SUES-200 dataset, SpikeViMFormer-T, which has the same 9.78M parameter count as the Baseline-T, achieves improvements in 13 metrics. For instance, in the drone$\rightarrow$satellite scenario, it achieves R@1 of 75.80\% (+1.25\%) and  AP of 80.13\% (+1.2\%) at 150m, R@1 of 84.53\% (+0.71\%) and  AP of 87.44\% (+0.59\%) at 200m, R@1 of 89.75\% (+1.43\%) and  AP of 91.63\% (+1.03\%) at 250m, and  R@1 of 91.10\% (+1.05\%) and AP of 92.70\% (+0.78\%) at 300m. Additionally, SpikeViMFormer-S also outperforms the Baseline-S with the same parameter count, which achieves improvements in 13 metrics. Moreover, on the University-1652 dataset, SpikeViMFormer maintains energy consumption comparable with the baseline, which further validates the effectiveness.

Secondly, ablations are performed on individual components to assess their respective contributions. The results are reported in Table \ref{tab:university-ab}.

\textbf{Effectiveness of SSA}.
The SSA block is validated in SpikeViMFormer-T and SpikeViMFormer-S. The results demonstrate that the incorporation of SSA consistently improves performance in both drone$\rightarrow$satellite and satellite$\rightarrow$drone scenarios. For instance, in SpikeViMFormer-T, the incorporation of SSA raises R@1 from 76.59\% to 77.85\% and AP from 80.39\% to 81.31\% in drone$\rightarrow$satellite scenario. Similarly, in SpikeViMFormer-S, SSA improves R@1 by nearly 4\% and AP by more than 3\% in drone$\rightarrow$satellite scenario. These improvements are attributed to the proposed SSA block, which enhances feature discriminability and learns critical information in the DVGL task.

\textbf{Effectiveness of SHS}.
As shown in Table \ref{tab:university-ab}, the SHS block achieves performance improvements in both versions. For example, in SpikeViMFormer-T, SHS further improves R@1 in drone$\rightarrow$satellite scenario from 76.59\% to 82.13\% and AP from 80.39\% to 85.01\%. In SpikeViMFormer-S, the same component raises R@1 from 78.65\% to 85.87\% and AP from 82.12\% to 88.23\%. The SHS block can effectively enhance semantic consistency by capturing long-range dependencies. Furthermore, the combination of SSA and SHS leads to the overall performance improvement. For instance, SpikeViMFormer-T with both SSA and SHS achieves R@1 of 84.31\% and AP of 86.89\% in drone$\rightarrow$satellite scenario, and R@1 of 92.43\% and AP of 84.39\% in satellite$\rightarrow$drone scenario, which demonstrates clear synergistic effects between SSA block and SHS block.

\textbf{Effectiveness of HRAL.}
As detailed in Table \ref{tab:university-ab}, the incorporation of the HRAL strategy leads to further performance improvements in most metrics, despite a slight decline in satellite$\rightarrow$drone scenario. Specifically, in SpikeViMFormer-T already equipped with SSA and SHS, HRAL raises drone$\rightarrow$satellite scenario R@1 from 84.31\% to 86.10\% and AP from 86.89\% to 88.32\%. Although a minor decrease is observed in satellite$\rightarrow$drone R@1 (from 92.43\% to 91.58\%), the AP still improves from 84.39\% to 85.40\%. In SpikeViMFormer-S, HRAL makes performance higher, which achieves 88.03\% in R@1 and 89.98\% in AP for drone$\rightarrow$satellite scenario from 87.21\% and 89.35\%, respectively. A similar slight drop occurs in R@1 (from 93.29\% to 92.72\%), while AP improves from 86.09\% to 87.10\% in satellite$\rightarrow$drone scenario. These results demonstrate that HRAL contributes to the overall accuracy of SpikeViMFormer, and confirm its value in cross-view learning.

\begin{table*}[t]
\centering
\caption{Comparison of SpikeViMFormer with state-of-the-art SNNs under the RandomKeepPatches process on the drone$\rightarrow$satellite of the University-1652.}
\label{RandomKeepPatches}
\resizebox{\textwidth}{!}{
\begin{tabular}{clcccccccccccc}
\hline
 \multirow{2}{*}{Setting} &Ratio (\%)     &  & \multicolumn{2}{c}{SpikeViMFormer-S} &  & \multicolumn{2}{c}{Meta-SpikeFormer} &  & \multicolumn{2}{c}{E-SpikeFormer-1} &  &  \multicolumn{2}{c}{E-SpikeFormer-2} 
\\ \cline{4-5} \cline{7-8} \cline{10-11} \cline{13-14} 
&Patches    &  & R@1      & AP       &  & R@1     & AP           &  & R@1    & AP       &  & R@1     & AP     
\\ \hline
1&$R$=20,$P$=5     &  &   \textbf{23.34$_{-64.69}$}    & \textbf {28.26$_{-61.72}$}     &  &   15.97$_{-62.97}$  &   20.61$_{-61.46}$     &  &      {19.67$_{-61.83}$}  &   {24.70$_{-59.84}$} &  &      21.84$_{-60.65}$   &   26.79$_{-58.69}$
\\
2&$R$=40,$P$=5     &  &  \textbf{49.31$_{-38.72}$}  & \textbf {54.57$_{-35.41}$}   &  &   36.26$_{-42.68}$  &  42.29$_{-39.78}$       &  &     {40.48$_{-41.02}$}  &   {46.38$_{-38.16}$}   &  &    43.33$_{-39.61}$    &  49.03$_{-36.45}$
\\
3&$R$=60,$P$=5     &  & \textbf{67.20$_{-20.83}$}  & \textbf{71.54$_{-18.44}$}    &  &    51.30$_{-27.64}$   & 56.93$_{-25.14}$       &  &  {55.20$_{-26.30}$} & {60.46$_{-24.08}$}  &  &  58.36$_{-24.13}$   &     63.41$_{-24.07}$ 
\\
4&$R$=80,$P$=5     &  &\textbf{74.39$_{-13.64}$}   &\textbf{78.02$_{-11.96}$}   &  &   59.42$_{-19.52}$  &   64.69$_{-17.11}$      &  &    {62.34$_{-19.16}$}  & {67.42$_{-17.12}$}   &  &    65.18$_{-14.31}$  &      70.00$_{-15.48}$
\\
5&$R$=100,$P$=5    &  &\textbf{78.93$_{-9.100}$}   &\textbf{82.11$_{-7.780}$}   &  &   65.19$_{-13.75}$    &  69.95$_{-12.12}$     &  & {66.96$_{-14.54}$}   & {71.61$_{-12.93}$} &  &  69.81$_{-12.68}$    &     74.19$_{-11.29}$
\\
\hline
6&$R$=20,$P$=10     &  &\textbf{25.88$_{-62.15}$}    &\textbf{31.25$_{-58.73}$}     &  &   18.53$_{-60.41}$  &   23.63$_{-58.44}$        &  &   {21.11$_{-60.39}$}   & {26.37$_{-58.17}$} &  &      23.18$_{-59.31}$   &   28.42$_{-57.06}$  
\\
7&$R$=40,$P$=10     &  &\textbf{49.75$_{-38.28}$}  &\textbf{55.22$_{-34.76}$}   &  &   38.84$_{-40.05}$  & 44.79$_{-37.28}$  &  &     {40.06$_{-41.44}$}  &  {45.96$_{-38.58}$}    &  &    43.37$_{-39.12}$    &  49.11$_{-36.37}$    
\\
8&$R$=60,$P$=10     &  &\textbf{64.52$_{-23.51}$}  &\textbf{69.12$_{-20.86}$}    &  &    50.38$_{-28.56}$   &  56.12$_{-25.95}$       &  &  {54.03$_{-27.47}$} &  {59.55$_{-24.99}$}  &  &   56.85$_{-25.64}$   &     62.24$_{-23.24}$ 
\\
9&$R$=80,$P$=10     &  &\textbf{73.31$_{-14.72}$}   &\textbf{77.18$_{-12.80}$}   &  &   60.67$_{-18.27}$  &   65.81$_{-16.26}$     &  &    {62.68$_{-18.82}$}  &  {67.62$_{-16.92}$}   &  &    64.60$_{-17.89}$  &      69.31$_{-16.17}$ 
\\
10&$R$=100,$P$=10    &  &\textbf{77.55$_{-10.48}$}   &\textbf {80.94$_{-9.040}$}   &  &   64.83$_{-14.11}$    &  69.67$_{-12.40}$    &  &     {66.83$_{-14.67}$}   & {71.52$_{-13.02}$}  &  &  69.23$_{-13.26}$    &     73.73$_{-11.75}$
\\
\hline
\label{tab:PR}
\end{tabular}}
\end{table*}

\subsection{Critical Information and Long-Range Dependencies}
To further evaluate the ability of SpikeViMFormer to enhance critical information under the condition of sparse activations, as well as its capability of learning long-range dependencies, a novel processing strategy RandomKeepPatches is proposed that applied in the inference stage. Unlike traditional padding or mirroring-based augmentation methods \cite{shen2023mccg,chen2024sdpl}, it constructs more extreme experimental conditions by sparsification of the input images.

\begin{figure}[t]
  \centering
  \includegraphics[width=3.4in]{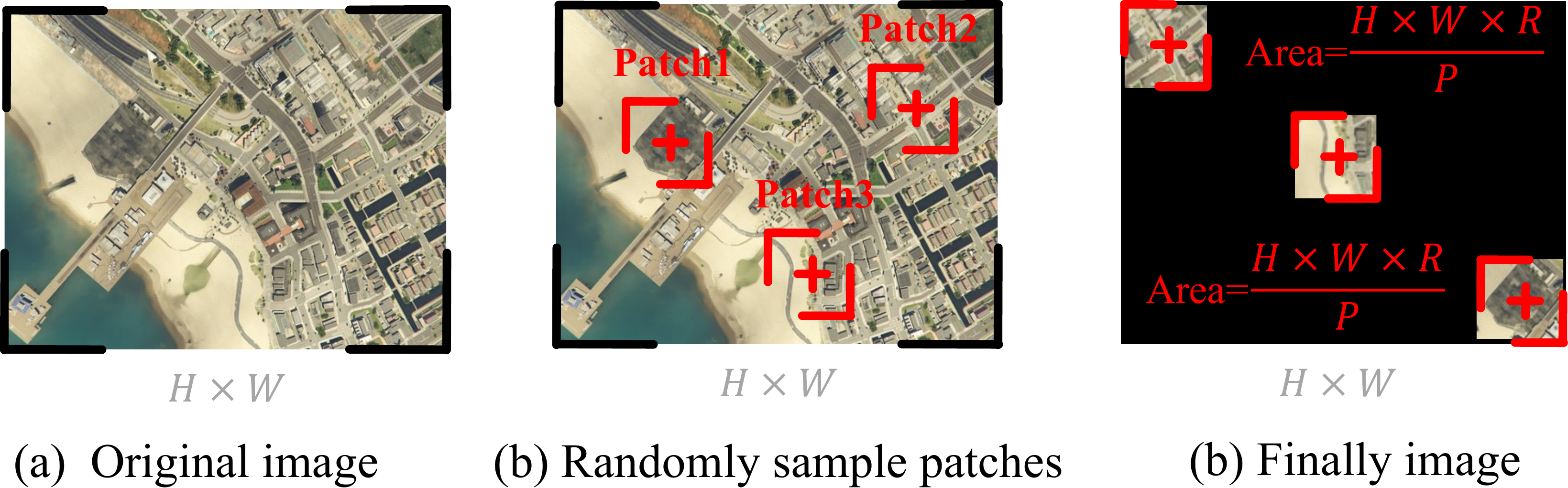}
  \caption{\textbf{RandomKeepPatches Strategy}. 
(a) The input image during inference stage with spatial resolution $H \times W$. 
(b) A set of patches is randomly sampled from the input image. 
(c) The final input of the sparse activation is constructed by randomly placing $P$ patches, each with an area of $\tfrac{H \times W \times R}{P}$, where $R$ denotes the predefined retention ratio of the total image area.}
  \label{fig6}
  
\end{figure}

\textbf{RandomKeepPatches Strategy}. As shown in Fig. \ref{fig6}, the total area to be preserved is first calculated according to a predefined retention ratio $R$ and then  this area is divided into small $P$ patch regions. They are placed at random positions. Only the original content within these patches is copied to the new image, while all remaining unselected regions are set to 0. As a result, the generated samples exhibit highly sparse and fragmented characteristics, where the original continuous contextual information is deliberately disrupted and only a few scattered local regions remain valid sources of information. Under such conditions, completing the DVGL task requires the testee to rely on limited critical information and simultaneously integrate spatially separated cues. It should be emphasized that during the random placement of patches, both repetition and overlap are allowed. Consequently, even when the predefined retention ratio $R$ is set to 100, the constructed image is almost never identical to the original one.

\textbf{Results in RandomKeepPatches}. 
To evaluate the performance of SpikeViMFormer under such extreme conditions, it is compared with SOTA SNNs using the RandomKeepPatches strategy. The retention ratio $R$ ranges from $20$ to $100$, and the number of preserved patches $P$ is set to $5$ and $10$. Specifically, $P=5$ retains fewer regions with low overlap risk to emphasize the importance of critical information, whereas $P=10$ retains more regions but increases the risk of overlap, which may result in confusion. This design enables an evaluation of how critical information and long-range dependencies can be learned under varying degrees of sparsification.

\begin{figure}[t]
  \centering
  \includegraphics[width=3.5in]{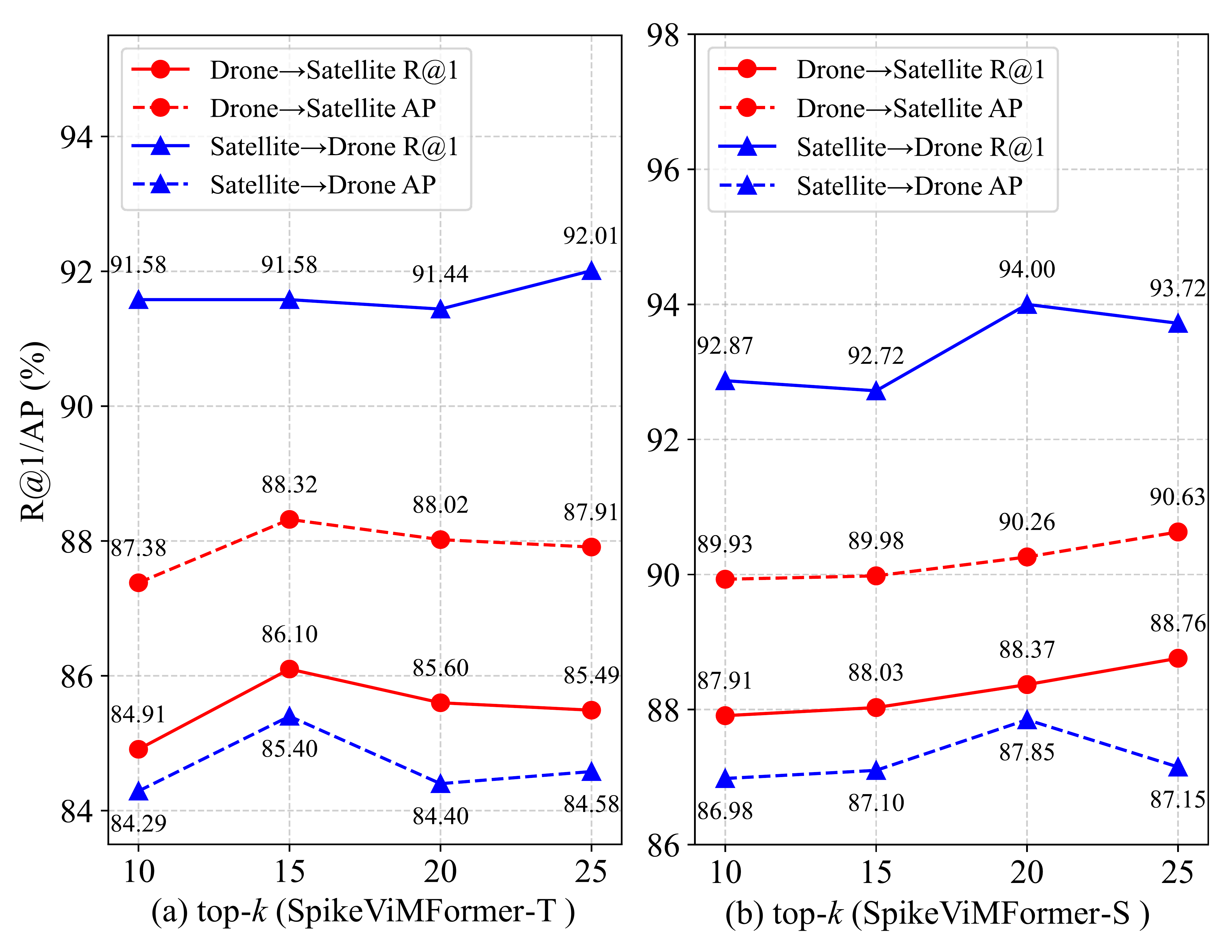}
  \caption{\textbf{Hyper-parameter Analysis}. Evaluation of hyper-parameter $k$. The results are test on University-1652.
R@1 and AP are reported.}
  \label{fig7}
\end{figure}

\begin{table}[ht]
  \centering
  \footnotesize
  \caption{Performance of SpikeViMFormer-S with different $\lambda_1$ and $\lambda_2$ on University-1652.}
\resizebox{\columnwidth}{!}{ 
  \setlength{\tabcolsep}{3pt}
  \begin{tabular}{ccccc|ccccc}
  \hline
    \multirow{2}{*}{$\lambda_1$} & \multicolumn{2}{c}{Drone$\to$Satellite} & \multicolumn{2}{c|}{Satellite$\to$Drone} &\multirow{2}{*}{$\lambda_2$} & \multicolumn{2}{c}{Drone$\to$Satellite} & \multicolumn{2}{c}{Satellite$\to$Drone} \\
    \cline{2-5} \cline{7-10}
     & R@1 & AP & R@1 & AP && R@1 & AP & R@1 & AP \\\hline
0.2 & 86.95 & 89.00 & 91.29 & 86.57 &0.34 &87.57 &89.48 &92.19 &87.02\\
0.4 & 87.52 & 89.45 & 91.72 & 86.65 &0.44 &87.81 &89.66 &92.57 &\textbf{87.23}\\
0.6 & \textbf{88.01} & \textbf{89.98} & \textbf{92.72} & \textbf{87.10} &0.54 &\textbf{88.01} &\textbf{89.98} &\textbf{92.72} &87.10\\
0.8 & 87.52 & 89.35 & 91.90 & 86.72 &0.64 &87.27 &89.22 &91.87 &86.88\\
1.0 & 87.58 & 90.51 & 91.15 & 86.04 &0.74 &87.39 &89.32 &91.57 &86.42\\\hline
  \end{tabular}}
  \label{tab:lamuta1}
\end{table}

SpikeViMFormer is  evaluated in the drone$\rightarrow$satellite scenario on the University-1652 dataset. As shown in Table~\ref{RandomKeepPatches}, SpikeViMFormer consistently outperforms SOTA SNNs in all ten settings. It achieves higher R@1 and AP. Notably, while the largest performance degradation occurs in the most challenging settings 1 and 6, SpikeViMFormer exhibits the smallest degradation in all other settings compared with SOTA SNNs. SpikeViMFormer not only maintains robustness under extreme sparsity, but also better preserves critical information and long-range dependencies.

\begin{figure*}[t]
  \centering
  \includegraphics[width=7.2in]{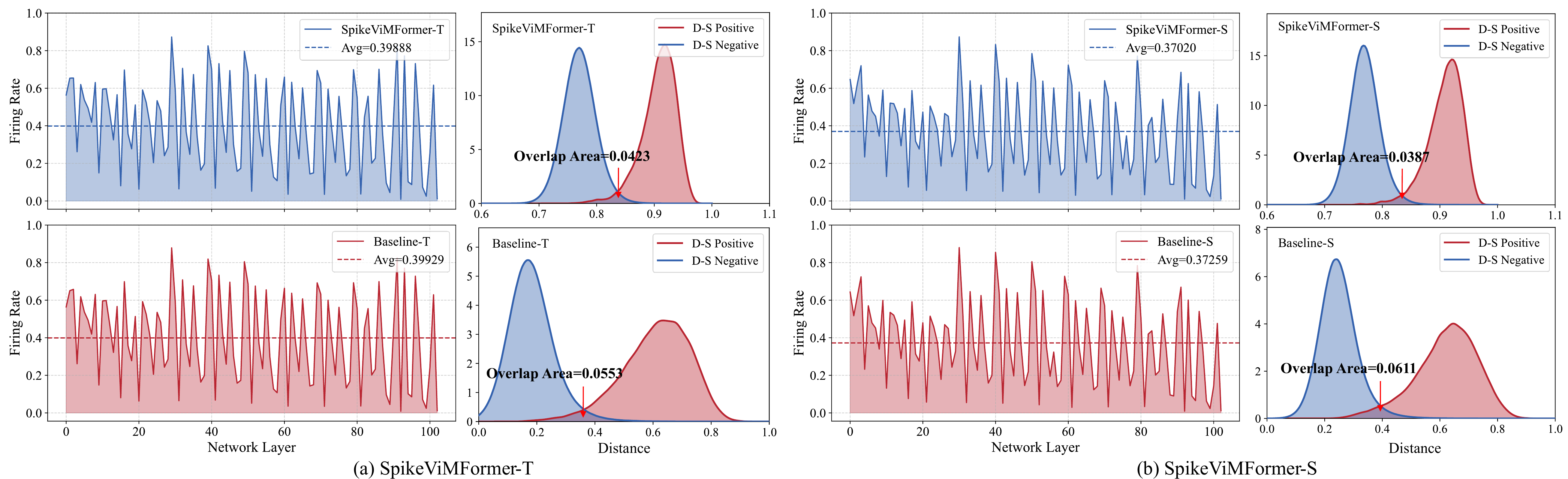}
  \caption{\textbf{Spike Firing Rate and Similarity Distribution}. (a) The first row corresponds to SpikeViMFormer-T, and the second row shows the results of the corresponding baseline-T. The left part shows the spike firing rates across layers, while the right part illustrates the similarity distributions of positive and negative samples. The red-shaded area in the similarity plots indicates the overlapping region between the positive and negative distributions. (b) SpikeViMFormer-S, with the same presentation as in (a).}
  \label{fig5}
\end{figure*}

\begin{figure*}[t]
  \centering
  \includegraphics[width=7.2in]{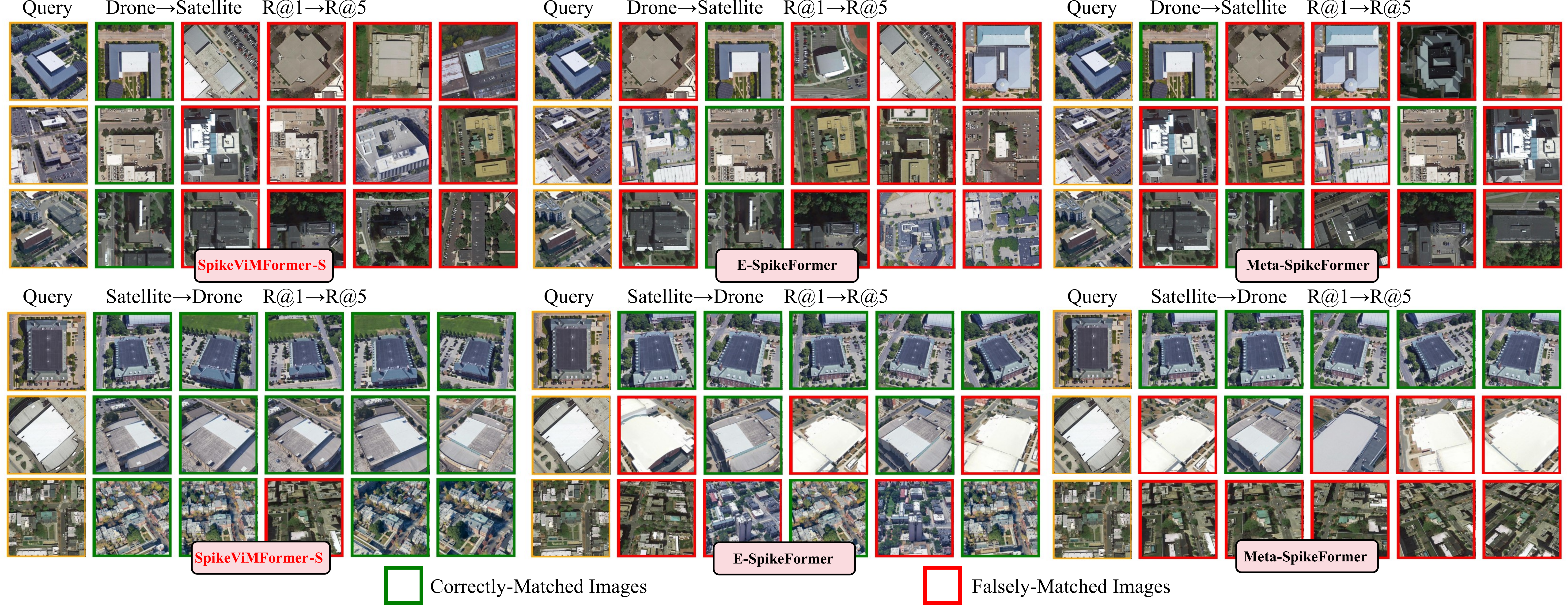}
  \caption{\textbf{Retrieval Results}. Retrieval examples of SpikeViMFormer-S, E-SpikeFormer, and Meta-SpikeFormer on the University-1652 dataset. R@1$\rightarrow$R@5 results are shown for satellite$\rightarrow$drone and drone$\rightarrow$satellite scenarios. Green boxes represent correct matches and red boxes
represent errors.}
  \label{fig8}
\end{figure*}

\subsection{Hyper-parameter Analysis}
\textbf{Analysis of $\lambda_1$ and $\lambda_2$}. As shown in Table~\ref{tab:lamuta1}, the performance of SpikeViMFormer-S remains relatively stable in different hyper-parameters $\lambda_1$ and $\lambda_2$ in Eq.~(\ref{eq34}), which demonstrates the robustness of the proposed SNN framework to these parameters. When $\lambda_1 = 0.6$ and $\lambda_2 = 0.54$, SpikeViMFormer-S achieves the best results on both drone$\to$satellite and satellite$\to$drone scenarios. Therefore, these values are adopted in this paper, while further fine-tuning may achieve even better performance.

\textbf{Analysis of $k$}. In Fig. \ref{fig7}, the empirical analysis is illustrated on the hyper-parameter $k$ used in top-$k$ neighbor list of the proposed HRAL strategy. It shows that the performance metrics of both SpikeViMFormer-T and SpikeViMFormer-S remain largely consistent in $k$ ranging from 10 to 25, which indicates strong robustness with respect to the choice of neighborhood size. For SpikeViMFormer-T, increasing $k$ does not lead to noticeable improvements due to its limited representational capacity. The incorporation of additional neighbors may introduce noise rather than beneficial context. In contrast, SpikeViMFormer-S exhibits a slight improvement in R@1 around $k$=20, indicating that deeper architectures can more effectively leverage information from a moderately enlarged neighborhood. Stable performance in all evaluation metrics confirms the strong generalization capability of the HRAL without requiring extensive tuning. Based on these results, $k$ is recommended in the range of [15,20] to achieve an optimal balance between performance and computational practicality.

\subsection{Visualization}

\textbf{Spike Firing Rate and Similarity Distribution}. The spike firing rate \( R_f \) is closely related to the energy consumption of the SNN network, as described in Eq.(\ref{eq37}). To further evaluate the energy consumption and discriminability of SpikeViMFormer compared to the baseline, we visualize the spike firing rate and the similarity distribution of positive and negative samples. As shown in Fig. \ref{fig5}, SpikeViMFormer maintains similar average spike firing rates in both versions compared to their respective baselines and even shows a slight decrease. Additionally, the similarity distribution demonstrates a reduction in the overlap area of positive and negative samples for SpikeViMFormer compared to the baseline, which facilitates clearer discrimination. This phenomenon further validates the conclusions reported in Tables \ref{tab:baseline} and \ref{tab:SUES-200baseline}. SpikeViMFormer does not enhance representation capabilities by increasing the spike firing rate, but rather achieves cross-view alignment by  mining critical information and long-range dependencies.

\textbf{Retrieval Visualization}. To further validate the advantages of SpikeViMFormer against SOTA SNNs, a real-world simulation evaluation is conducted at six randomly selected locations from the University-1652 dataset, for drone$\rightarrow$satellite and satellite$\rightarrow$drone scenarios, each with three locations. As shown in Fig. \ref{fig8}, for drone$\rightarrow$satellite queries where only one corresponding satellite image exists, the ground truth is expected to be retrieved at R@1. SpikeViMFormer-S successfully retrieves the correct match at R@1 at all three locations that demonstrates satisfied accuracy and robustness. In contrast, E-SpikeFormer achieves correct retrieval only at R@2, while Meta-SpikeFormer exhibits unstable performance retrieving correctly at R@1 in only one location, and at R@2 and R@4 in the other two locations, respectively. Moreover, in satellite$\rightarrow$drone scenario, although all three methods perform well in simple scenarios, SpikeViMFormer-S shows superior discrimination capability in challenging regions, which has highlighted its stronger performance in the real-world.

\section{Conclusion}\label{conclusions}
This work addresses the challenges posed by resource-constrained drones as well as the substantial energy demands in DVGL. Compared with ANNs, SNNs offer a more energy-efficient paradigm for DVGL. Nevertheless, their sparse activation tends to cause the loss of critical information and limits the ability to model long-range dependencies. Therefore, we propose SpikeViMFormer, the first hardware-compatible and high-performance SNN framework for DVGL. In this framework, SpikeViMFormer integrates the SSA and SHS blocks in an end-to-end manner to selectively enhance critical information and capture long-range dependencies. Additionally, auxiliary blocks are used only during the training stage to ensure lightweight inference, and the HRAL strategy is proposed to enhance backbone performance through neighbor information and cross-batch consistency. Extensive experiments demonstrate that SpikeViMFormer outperforms SOTA SNNs and remains competitive with advanced ANNs. It also offers significant potential advantages in energy efficiency and hardware deployment. In future work, we will further explore higher-performance SNN frameworks for DVGL and their deployment on neuromorphic hardware.

\bibliographystyle{IEEEtran}
\footnotesize
\bibliography{main}

@String(AAAI = {AAAI})

@article{xia2024enhancing,
  title={Enhancing cross-view geo-localization with domain alignment and scene consistency},
  author={Xia, Panwang and Wan, Yi and Zheng, Zhi and Zhang, Yongjun and Deng, Jiwei},
  journal={{IEEE} Trans. Circuits Syst. Video Technol.},
  year={2024},
  volume={},
  number={},
  pages={1-12},
  publisher={IEEE}
}

@article{tian2021uav,
  title={{UAV}-satellite view synthesis for cross-view geo-localization},
  author={Tian, Xiaoyang and Shao, Jie and Ouyang, Deqiang and Shen, Heng Tao},
  journal={{IEEE} Trans. Circuits Syst. Video Technol.},
  volume={32},
  number={7},
  pages={4804-4815},
  year={2021},
  publisher={IEEE}
}

@article{dai2021transformer,
  title={A transformer-based feature segmentation and region alignment method for {UAV}-view geo-localization},
  author={Dai, Ming and Hu, Jianhong and Zhuang, Jiedong and Zheng, Enhui},
  journal={{IEEE} Trans. Circuits Syst. Video Technol.},
  volume={32},
  number={7},
  pages={4376-4389},
  year={2021},
  publisher={IEEE}
}

@article{du2024ccr,
  title={{CCR}: A counterfactual causal reasoning-based method for cross-view geo-localization},
  author={Du, Haolin and He, Jingfei and Zhao, Yuanqing},
  journal={{IEEE} Trans. Circuits Syst. Video Technol.},
  year={2024},
  volume={34},
  number={11},
  pages={11630-11643},
}

@article{chen2024multi,
  title={Multilevel Embedding and Alignment Network With Consistency and Invariance Learning for Cross-View Geo-Localization}, 
  author={Chen, Zhongwei and Yang, Zhao-Xu and Rong, Hai-Jun},
  journal={{IEEE} Trans. Geosci. Remote Sens.}, 
  year={2025},
  volume={63},
  number={},
  pages={1-15},
}

@inproceedings{qin2025must,
  title={MUST: The First Dataset and Unified Framework for Multispectral UAV Single Object Tracking},
  author={Qin, Haolin and Xu, Tingfa and Li, Tianhao and Chen, Zhenxiang and Feng, Tao and Li, Jianan},
  booktitle={Proc. IEEE Conf. Comput. Vis. Pattern Recognit.},
  pages={16882--16891},
  year={2025}
}

@InProceedings{Liu_Multi,
    author    = {Liu, Shuai and Li, Xin and Lu, Huchuan and He, You},
    title     = {Multi-Object Tracking Meets Moving UAV},
    booktitle = {Proc. IEEE Conf. Comput. Vis. Pattern Recognit.},
    year      = {2022},
    pages     = {8876-8885}
}

@inproceedings{zheng2020university,
author = {Zheng, Zhedong and Wei, Yunchao and Yang, Yi},
title = {University-1652: {A} Multi-view Multi-source Benchmark for Drone-based Geo-localization},
year = {2020},
booktitle = {Proc. ACM Int. Conf. Multimedia},
pages = {1395–1403},
}

@article{zhu2023sues,
  title={{SUES}-200: {A} multi-height multi-scene cross-view image benchmark across drone and satellite},
  author={Zhu, Runzhe and Yin, Ling and Yang, Mingze and Wu, Fei and Yang, Yuncheng and Hu, Wenbo},
  journal={{IEEE} Trans. Circuits Syst. Video Technol.},
  volume={33},
  number={9},
  pages={4825-4839},
  year={2023},
  publisher={IEEE}
}

@article{chen2024sdpl,
  title={{SDPL}: {S}hifting-dense partition learning for {UAV}-view geo-localization},
  author={Chen, Quan and Wang, Tingyu and Yang, Zihao and Li, Haoran and Lu, Rongfeng and Sun, Yaoqi and Zheng, Bolun and Yan, Chenggang},
  journal={{IEEE} Trans. Circuits Syst. Video Technol.}, 
  volume={34},
  number={11},
  pages={11810-11824},
  year={2024},
  publisher={IEEE}
}

@article{ge2024multibranch,
  title={Multibranch joint representation learning based on information fusion strategy for cross-view geo-localization},
  author={Ge, Fawei and Zhang, Yunzhou and Liu, Yixiu and Wang, Guiyuan and Coleman, Sonya and Kerr, Dermot and Wang, Li},
  journal={{IEEE} Trans. Geosci. Remote Sens.},
  volume={62},
  pages={1-16},
  year={2024},
  publisher={IEEE}
}

@article{chen2025without,
  title={Without Paired Labeled Data: End-to-End Self-Supervised Learning for Drone-view Geo-Localization},
  author={Chen, Zhongwei and Yang, Zhao-Xu and Rong, Hai-Jun},
  journal={arXiv preprint arXiv:2502.11381},
  year={2025}
}

@inproceedings{liu2022convnet,
  title={A convNet for the 2020s},
  author={Liu, Zhuang and Mao, Hanzi and Wu, Chao-Yuan and Feichtenhofer, Christoph and Darrell, Trevor and Xie, Saining},
  booktitle = {Proc. IEEE Conf. Comput. Vis. Pattern Recognit.},
  year= {2022},
  pages = {11976-11986}

}

@inproceedings{liu2022swin,
  title={Swin transformer v2: {S}caling up capacity and resolution},
  author={Liu, Ze and Hu, Han and Lin, Yutong and Yao, Zhuliang and Xie, Zhenda and Wei, Yixuan and Ning, Jia and Cao, Yue and Zhang, Zheng and Dong, Li and others},
    booktitle = {Proc. IEEE Conf. Comput. Vis. Pattern Recognit.},
    year      = {2022},
    pages     = {12009-12019}
}

@article{pei2019towards,
  title={Towards artificial general intelligence with hybrid Tianjic chip architecture},
  author={Pei, Jing and Deng, Lei and Song, Sen and Zhao, Mingguo and Zhang, Youhui and Wu, Shuang and Wang, Guanrui and Zou, Zhe and Wu, Zhenzhi and He, Wei and others},
  journal={Nature},
  volume={572},
  number={7767},
  pages={106--111},
  year={2019}
}

@article{yao2023spike,
  title={Spike-driven transformer},
  author={Yao, Man and Hu, Jiakui and Zhou, Zhaokun and Yuan, Li and Tian, Yonghong and Xu, Bo and Li, Guoqi},
  journal={in Proc. Adv. Neural Inf. Process. Syst.},
  volume={36},
  pages={64043--64058},
  year={2023}
}

@article{yao2023attention,
  title={Attention spiking neural networks},
  author={Yao, Man and Zhao, Guangshe and Zhang, Hengyu and Hu, Yifan and Deng, Lei and Tian, Yonghong and Xu, Bo and Li, Guoqi},
  journal={IEEE Trans. Pattern Anal. Mach. Intell.	},
  volume={45},
  number={8},
  pages={9393--9410},
  year={2023},
  publisher={IEEE}
}

@article{deng2020model,
  title={Model compression and hardware acceleration for neural networks: A comprehensive survey},
  author={Deng, Lei and Li, Guoqi and Han, Song and Shi, Luping and Xie, Yuan},
  journal={Proceedings of the IEEE},
  volume={108},
  number={4},
  pages={485--532},
  year={2020},
  publisher={IEEE}
}

@article{tavanaei2019deep,
  title={Deep learning in spiking neural networks},
  author={Tavanaei, Amirhossein and Ghodrati, Masoud and Kheradpisheh, Saeed Reza and Masquelier, Timoth{\'e}e and Maida, Anthony},
  journal={Neural networks},
  volume={111},
  pages={47--63},
  year={2019},
  publisher={Elsevier}
}

@inproceedings{deng2009imagenet,
  title={Imagenet: A large-scale hierarchical image database},
  author={Deng, Jia and Dong, Wei and Socher, Richard and Li, Li-Jia and Li, Kai and Fei-Fei, Li},
  booktitle={Proc. IEEE Conf. Comput. Vis. Pattern Recognit.},
  pages={248-255},
  year={2009}
}

@inproceedings{meng2023towards,
  title={Towards memory-and time-efficient backpropagation for training spiking neural networks},
  author={Meng, Qingyan and Xiao, Mingqing and Yan, Shen and Wang, Yisen and Lin, Zhouchen and Luo, Zhi-Quan},
  booktitle={Proc. IEEE Int. Conf. Comput. Vis.},
  pages={6166--6176},
  year={2023}
}

@inproceedings{
yao2024spikedriven,
title={Spike-driven Transformer V2: Meta Spiking Neural Network Architecture Inspiring the Design of Next-generation Neuromorphic Chips},
author={Man Yao and JiaKui Hu and Tianxiang Hu and Yifan Xu and Zhaokun Zhou and Yonghong Tian and Bo XU and Guoqi Li},
booktitle={Proc. Int. Conf. Learn. Represent.},
year={2024}
}

@ARTICLE{10848017,
  author={Yao, Man and Qiu, Xuerui and Hu, Tianxiang and Hu, Jiakui and Chou, Yuhong and Tian, Keyu and Liao, Jianxing and Leng, Luziwei and Xu, Bo and Li, Guoqi},
  journal={IEEE Trans. Pattern Anal. Mach. Intell.}, 
  title={Scaling Spike-Driven Transformer With Efficient Spike Firing Approximation Training}, 
  year={2025},
  volume={47},
  number={4},
  pages={2973-2990},
  doi={10.1109/TPAMI.2025.3530246}}

@article{davies2018loihi,
  title={Loihi: A neuromorphic manycore processor with on-chip learning},
  author={Davies, Mike and Srinivasa, Narayan and Lin, Tsung-Han and Chinya, Gautham and Cao, Yongqiang and Choday, Sri Harsha and Dimou, Georgios and Joshi, Prasad and Imam, Nabil and Jain, Shweta and others},
  journal={Ieee Micro},
  volume={38},
  number={1},
  pages={82--99},
  year={2018},
  publisher={IEEE}
}

@article{yao2024spike,
  title={Spike-based dynamic computing with asynchronous sensing-computing neuromorphic chip},
  author={Yao, Man and Richter, Ole and Zhao, Guangshe and Qiao, Ning and Xing, Yannan and Wang, Dingheng and Hu, Tianxiang and Fang, Wei and Demirci, Tugba and De Marchi, Michele and others},
  journal={Nat. Commun.	},
  volume={15},
  number={1},
  pages={4464},
  year={2024},
  publisher={Nature Publishing Group UK London}
}

@article{merolla2014million,
  title={A million spiking-neuron integrated circuit with a scalable communication network and interface},
  author={Merolla, Paul A and Arthur, John V and Alvarez-Icaza, Rodrigo and Cassidy, Andrew S and Sawada, Jun and Akopyan, Filipp and Jackson, Bryan L and Imam, Nabil and Guo, Chen and Nakamura, Yutaka and others},
  journal={Science},
  volume={345},
  number={6197},
  pages={668--673},
  year={2014},
  publisher={American Association for the Advancement of Science}
}

@article{ding2020practical,
  title={A practical cross-view image matching method between {UAV} and satellite for {UAV}-based geo-localization},
  author={Ding, Lirong and Zhou, Ji and Meng, Lingxuan and Long, Zhiyong},
  journal={Remote Sensing},
  volume={13},
  number={1-20},
  pages={47},
  year={2020},
  publisher={MDPI}
}

@article{wang2021each,
  title={Each part matters: {L}ocal patterns facilitate cross-view geo-localization},
  author={Wang, Tingyu and Zheng, Zhedong and Yan, Chenggang and Zhang, Jiyong and Sun, Yaoqi and Zheng, Bolun and Yang, Yi},
  journal={IEEE Trans. Circuits Syst. Video Technol.},
  volume={32},
  number={2},
  pages={867-879},
  year={2021},
  publisher={IEEE}
}

@article{ge2024multi,
  title={Multilevel feedback joint representation learning network based on adaptive area elimination for cross-view geo-localization},
  author={Ge, Fawei and Zhang, Yunzhou and Wang, Li and Liu, Wei and Liu, Yixiu and Coleman, Sonya and Kerr, Dermot},
  journal={IEEE Trans. Geosci. Remote Sens.},
  year={2024},
  volume={62},
  number={},
  pages={1-15},
  publisher={IEEE}
}

@article{wu2024camp,
  title={{CAMP}: {A}cross-view geo-localization method using contrastive attributes mining and position-aware partitioning},
  author={Wu, Qiong and Wan, Yi and Zheng, Zhi and Zhang, Yongjun and Wang, Guangshuai and Zhao, Zhenyang},
  journal={IEEE Trans. Geosci. Remote Sens.},
  year={2024},
  volume={62},
  number={},
  pages={1-14},
}

@article{zhao2024transfg,
  title={{T}rans{FG}: {A} cross-view geo-localization of satellite and {UAV}s imagery pipeline using transformer-based feature aggregation and gradient guidance},
  author={Zhao, Hu and Ren, Keyan and Yue, Tianyi and Zhang, Chun and Yuan, Shuai},
  journal={IEEE Trans. Geosci. Remote Sens.},
  volume={62},
  number={},
  pages={1-12},
  year={2024},
  publisher={IEEE}
}

@article{wilson2021visual,
  title={Visual and object geo-localization: A comprehensive survey},
  author={Wilson, Daniel and Zhang, Xiaohan and Sultani, Waqas and Wshah, Safwan},
  journal={arXiv preprint arXiv:2112.15202},
  year={2021}
}

@article{mirza2014conditional,
  title={Conditional generative adversarial nets},
  author={Mirza, Mehdi and Osindero, Simon},
  journal={arXiv preprint arXiv:1411.1784},
  year={2014}
}

@article{schuman2022opportunities,
  title={Opportunities for neuromorphic computing algorithms and applications},
  author={Schuman, Catherine D and Kulkarni, Shruti R and Parsa, Maryam and Mitchell, J Parker and Date, Prasanna and Kay, Bill},
  journal={Nat. Comput. Sci.	},
  volume={2},
  number={1},
  pages={10--19},
  year={2022}
}

@article{maass1997networks,
  title={Networks of spiking neurons: the third generation of neural network models},
  author={Maass, Wolfgang},
  journal={Neural networks},
  volume={10},
  number={9},
  pages={1659--1671},
  year={1997},
  publisher={Elsevier}
}

@article{bi2001synaptic,
  title={Synaptic modification by correlated activity: Hebb's postulate revisited},
  author={Bi, Guo-qiang and Poo, Mu-ming},
  journal={Annu. Rev. Neurosci.},
  volume={24},
  number={1},
  pages={139--166},
  year={2001},
  publisher={Annual Reviews 4139 El Camino Way, PO Box 10139, Palo Alto, CA 94303-0139, USA}
}

@inproceedings{yao2021temporal,
  title={Temporal-wise attention spiking neural networks for event streams classification},
  author={Yao, Man and Gao, Huanhuan and Zhao, Guangshe and Wang, Dingheng and Lin, Yihan and Yang, Zhaoxu and Li, Guoqi},
  booktitle={Proc. IEEE Int. Conf. Comput. Vis.},
  pages={10221--10230},
  year={2021}
}

@article{hu2023fast,
  title={Fast-snn: Fast spiking neural network by converting quantized ann},
  author={Hu, Yangfan and Zheng, Qian and Jiang, Xudong and Pan, Gang},
  journal={IEEE Trans. Pattern Anal. Mach. Intell.	},
  volume={45},
  number={12},
  pages={14546--14562},
  year={2023},
  publisher={IEEE}
}

@article{eshraghian2023training,
  title={Training spiking neural networks using lessons from deep learning},
  author={Eshraghian, Jason K and Ward, Max and Neftci, Emre O and Wang, Xinxin and Lenz, Gregor and Dwivedi, Girish and Bennamoun, Mohammed and Jeong, Doo Seok and Lu, Wei D},
  journal={Proc. IEEE},
  volume={111},
  number={9},
  pages={1016--1054},
  year={2023},
  publisher={IEEE}
}

@inproceedings{luo2024integer,
  title={Integer-valued training and spike-driven inference spiking neural network for high-performance and energy-efficient object detection},
  author={Luo, Xinhao and Yao, Man and Chou, Yuhong and Xu, Bo and Li, Guoqi},
  booktitle={Proc. Eur. Conf. Comput. Vis.},
  pages={253--272},
  year={2024},
  organization={Springer}
}

@inproceedings{lei2025spike2former,
  title={Spike2former: Efficient spiking transformer for high-performance image segmentation},
  author={Lei, Zhenxin and Yao, Man and Hu, Jiakui and Luo, Xinhao and Lu, Yanye and Xu, Bo and Li, Guoqi},
  booktitle={Proc. AAAI Conf. Artif. Intell.},
  volume={39},
  number={2},
  pages={1364--1372},
  year={2025}
}

@article{benjamin2014neurogrid,
  title={Neurogrid: A mixed-analog-digital multichip system for large-scale neural simulations},
  author={Benjamin, Ben Varkey and Gao, Peiran and McQuinn, Emmett and Choudhary, Swadesh and Chandrasekaran, Anand R and Bussat, Jean-Marie and Alvarez-Icaza, Rodrigo and Arthur, John V and Merolla, Paul A and Boahen, Kwabena},
  journal={Proc. IEEE},
  volume={102},
  number={5},
  pages={699--716},
  year={2014},
  publisher={IEEE}
}

@article{hoppner2021spinnaker,
  title={The SpiNNaker 2 processing element architecture for hybrid digital neuromorphic computing},
  author={H{\"o}ppner, Sebastian and Yan, Yexin and Dixius, Andreas and Scholze, Stefan and Partzsch, Johannes and Stolba, Marco and Kelber, Florian and Vogginger, Bernhard and Neum{\"a}rker, Felix and Ellguth, Georg and others},
  journal={arXiv preprint arXiv:2103.08392},
  year={2021}
}

@article{chu2021conditional,
  title={Conditional positional encodings for vision transformers},
  author={Chu, Xiangxiang and Tian, Zhi and Zhang, Bo and Wang, Xinlong and Shen, Chunhua},
  journal={arXiv preprint arXiv:2102.10882},
  year={2021}
}

@article{vaswani2017attention,
  title={Attention is all you need},
  author={Vaswani, Ashish and Shazeer, Noam and Parmar, Niki and Uszkoreit, Jakob and Jones, Llion and Gomez, Aidan N and Kaiser, {\L}ukasz and Polosukhin, Illia},
  journal={in Proc. Adv. Neural Inf. Process. Syst.},
  volume={30},
  year={2017}
}

@article{dosovitskiy2020image,
  title={An image is worth 16x16 words: Transformers for image recognition at scale},
  author={Dosovitskiy, Alexey and Beyer, Lucas and Kolesnikov, Alexander and Weissenborn, Dirk and Zhai, Xiaohua and Unterthiner, Thomas and Dehghani, Mostafa and Minderer, Matthias and Heigold, Georg and Gelly, Sylvain and others},
  journal={arXiv preprint arXiv:2010.11929},
  year={2020}
}

@article{liu2024vmamba,
  title={Vmamba: Visual state space model},
  author={Liu, Yue and Tian, Yunjie and Zhao, Yuzhong and Yu, Hongtian and Xie, Lingxi and Wang, Yaowei and Ye, Qixiang and Jiao, Jianbin and Liu, Yunfan},
  journal={in Proc. Adv. Neural Inf. Process. Syst.},
  volume={37},
  pages={103031--103063},
  year={2024}
}

@inproceedings{lee2025efficientvim,
  title={Efficientvim: Efficient vision mamba with hidden state mixer based state space duality},
  author={Lee, Sanghyeok and Choi, Joonmyung and Kim, Hyunwoo J},
  booktitle={Proc. IEEE Conf. Comput. Vis. Pattern Recognit.},
  pages={14923--14933},
  year={2025}
}

@article{oord2018representation,
  title={Representation learning with contrastive predictive coding},
  author={Oord, Aaron van den and Li, Yazhe and Vinyals, Oriol},
  journal={arXiv preprint arXiv:1807.03748},
  year={2018}
}

@inproceedings{zhong2017re,
  title={Re-ranking person re-identification with k-reciprocal encoding},
  author={Zhong, Zhun and Zheng, Liang and Cao, Donglin and Li, Shaozi},
  booktitle={Proc. IEEE Conf. Comput. Vis. Pattern Recognit.},
  pages={1318--1327},
  year={2017}
}

@article{molchanov2016pruning,
  title={Pruning convolutional neural networks for resource efficient inference},
  author={Molchanov, Pavlo and Tyree, Stephen and Karras, Tero and Aila, Timo and Kautz, Jan},
  journal={arXiv preprint arXiv:1611.06440},
  year={2016}
}

@article{yin2021accurate,
  title={Accurate and efficient time-domain classification with adaptive spiking recurrent neural networks},
  author={Yin, Bojian and Corradi, Federico and Boht{\'e}, Sander M},
  journal={Nat. Mach. Intell.	},
  volume={3},
  number={10},
  pages={905--913},
  year={2021},
  publisher={Nature Publishing Group UK London}
}

@article{shen2023mccg,
  title={{MCCG}: {A} convNeXt-based multiple-classifier method for cross-view geo-localization},
  author={Shen, Tianrui and Wei, Yingmei and Kang, Lai and Wan, Shanshan and Yang, Yee-Hong},
  journal={IEEE Trans. Circuits Syst. Video Technol.},
  volume={34},
  number={3},
  pages={1456-1468},
  year={2023},
  publisher={IEEE}
}
\end{document}